\RequirePackage{scrlfile}
\PreventPackageFromLoading{subfig}
\documentclass[journal]{IEEEtran}

\usepackage{blindtext}
\usepackage{graphicx}
\usepackage{amssymb}
\usepackage{graphicx}
\usepackage{bbm}
\usepackage{mathtools}
\usepackage{mathrsfs}
\usepackage{diagbox}
\usepackage{booktabs}
\usepackage{multirow}
\usepackage[ruled]{algorithm2e}
\usepackage{bbding}
\usepackage[table,xcdraw]{xcolor}
\usepackage{epsfig}
\usepackage{cases} 
\usepackage{overpic}      
\usepackage{amsmath}
\usepackage{subfigure}
\usepackage{color}
\usepackage{hyperref}
\newcommand{\citet}[1]{\citeauthor{#1} \shortcite{#1}}

\makeatletter
\DeclareRobustCommand\onedot{\futurelet\@let@token\@onedot}
\def\@onedot{\ifx\@let@token.\else.\null\fi\xspace}

\makeatother

\begin{document}
%%%%%%%%% TITLE
\title{Dynamic Path-Controllable Deep Unfolding Network for Compressive Sensing}

\author{Jiechong~Song, Bin~Chen, Jian~Zhang,~\IEEEmembership{Member,~IEEE}
\thanks{Manuscript received August 30, 2022; revised January 5, 2023; accepted March 14, 2023. This work was supported in par by National Natural Science Foundation of China under Grant 61902009 and Shenzhen Research Project under Grant JCYJ20220531093215035. }% <-this % stops a space

\thanks{J.~Song, B.~Chen and J.~Zhang are with the School of Electronic and Computer Engineering, Peking University Shenzhen Graduate School, Shenzhen 518055. (\textit{Corresponding author: Jian Zhang.})

% , China and also with the Peng Cheng Laboratory, Shenzhen 518052, China.

}% <-this % stops a space
}

\markboth{2023}%
{Shell \MakeLowercase{\textit{et al.}}: Bare Demo of IEEEtran.cls for Journals}

\maketitle
%\thispagestyle{empty}

%%%%%%%%% ABSTRACT
\begin{abstract}
\noindent 
Deep unfolding network (DUN) that unfolds the optimization algorithm into a deep neural network has achieved great success in compressive sensing (CS) due to its good interpretability and high performance. Each stage in DUN corresponds to one iteration in optimization. At the test time, all the sampling images generally need to be processed by all stages, which comes at a price of computation burden and is also unnecessary for the images whose contents are easier to restore. In this paper, we focus on CS reconstruction and propose a novel \textbf{D}ynamic \textbf{P}ath-\textbf{C}ontrollable \textbf{D}eep \textbf{U}nfolding \textbf{N}etwork (\textbf{DPC-DUN}). DPC-DUN with our designed path-controllable selector can dynamically select a rapid and appropriate route for each image and is slimmable by regulating different performance-complexity tradeoffs. Extensive experiments show that our DPC-DUN is highly flexible and can provide excellent performance and dynamic adjustment to get a suitable tradeoff, thus addressing the main requirements to become appealing in practice. Codes are available at \url{https://github.com/songjiechong/DPC-DUN}.
% What is more, we make experiments on the compressive sensing reconstruction of MR images and also prove the effectiveness of our method.
% \vspace{-12pt}
\end{abstract}


\section{Introduction}
% ACM's consolidated article template, introduced in 2017, provides a
% consistent \LaTeX\ style for use across ACM publications, and
% incorporates accessibility and metadata-extraction functionality
% necessary for future Digital Library endeavors. Numerous ACM and
% SIG-specific \LaTeX\ templates have been examined, and their unique
% features incorporated into this single new template.

% If you are new to publishing with ACM, this document is a valuable
% guide to the process of preparing your work for publication. If you
% have published with ACM before, this document provides insight and
% instruction into more recent changes to the article template.

% The ``\verb|acmart|'' document class can be used to prepare articles
% for any ACM publication --- conference or journal, and for any stage
% of publication, from review to final ``camera-ready'' copy, to the
% author's own version, with {\itshape very} few changes to the source.

Compressive sensing (CS) is a considerable research interest from signal/image processing communities as a joint acquisition and reconstruction approach \cite{zhang2022bdu,candes2006robust}. The signal is first sampled and compressed simultaneously with random linear transformations. Then, the original signal can be reconstructed from far fewer measurements than that required by Nyquist sampling rate \cite{sankaranarayanan2012cs,liutkus2014imaging,zymnis2009compressed}. As CS can reduce the amount of information to be observed and processed while maintaining a reasonable reconstruction of the sparse or compressible signal, it has spawned many applications, including but not limited to medical imaging \cite{ szczykutowicz2010dual, zhang2022high}, 
% wireless remote monitoring \cite{zhang2012compressed}, 
image compression \cite{chen2019compressive}, single-pixel cameras \cite{duarte2008single,rousset2016adaptive}, remote sensing \cite{shen2013compressed}, image classification \cite{mou2022transcl}, and snapshot compressive imaging \cite{wu2021spatial,wu2021ddun}.

%%%%%%%%%%%%%%%%%%%%%%%%%%%%%%%%%%%%%%%%
\begin{figure*}[t]
\centering
\setlength{\abovecaptionskip}{3pt}
\setlength{\belowcaptionskip}{-0.cm}
\includegraphics[width=1.0\textwidth]{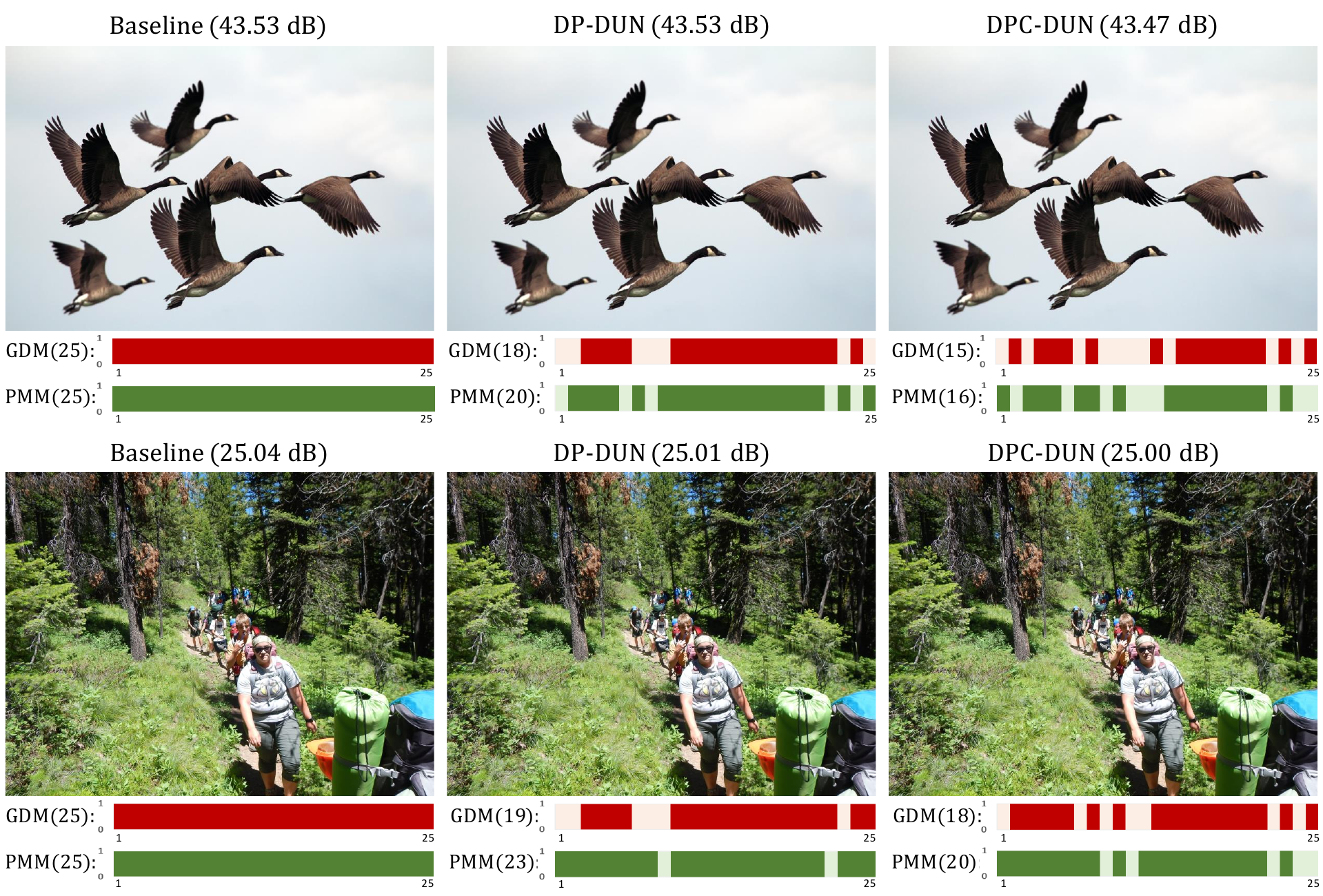} 
\vspace{-0.5cm}
\caption{Two subjective examples of the impact of the active module number on the reconstruction performance (PSNR). We compare the ``Baseline'' model without the selectors and our models (DP-DUN and DPC-DUN) with the selectors when ratio $=30\%$. While maintaining similar high performance, our methods can adaptively select an appropriate path and an optimal number of active modules for different images.}
\label{fig:intro}
\vspace{-0.5cm}
\end{figure*}
%%%%%%%%%%%%%%%%%%%%%%%%%%%%%%%%%%%%%%%%

Mathematically, a random linear measurement $\mathbf{y}\in\mathbb{R}^M$ can be formulated as $\mathbf{y} = \mathbf{\Phi}\mathbf{x}$, where $\mathbf{x}\in\mathbb{R}^N$ is the original signal and $\mathbf{\Phi}\in\mathbb{R}^{M \times N}$ is the measurement matrix with $M\ll N$. $\frac{M}{N}$ is the CS ratio (or sampling rate). Obviously, CS reconstruction is an ill-posed inverse problem. To obtain a reliable reconstruction, the conventional CS methods commonly solve an energy function as follows:
%%%%%%%%%%%%%%%%%%%%%%%%%%%%%%%%%%%%%%%%%%
\begin{align} \label{eq: opt}
\begin{split}
\setlength{\abovedisplayskip}{5pt}
\setlength{\belowdisplayskip}{6pt}
\underset{\mathbf{x}}{\arg\min} ~~\frac{1}{2} \left \|\mathbf{\Phi}\mathbf{x} - \mathbf{y} \right \|^2_2+ \lambda \mathcal{R}(\mathbf{x}),
% \left\|\boldsymbol{\Psi\boldsymbol{\mathbf{x}}} \right\|,
\end{split}
\end{align}
%%%%%%%%%%%%%%%%%%%%%%%%%%%%%%%%%%%%%%%%%%
where $\frac{1}{2} \left \|\mathbf{\Phi}\mathbf{x} - \mathbf{y} \right \|^2_2$ denotes the data-fidelity term and $\lambda \mathcal{R}(\mathbf{x})$ denotes the prior term with regularization parameter $\lambda$. 
% In this paper, we focus on CS reconstruction of natural and MR images, and our proposed framework can be also easily extended to video and other types of data. 
For traditional model-based methods \cite{kim2010compressed,Li2013AnEA,zhang2014group,zhang2014image,gao2015block,Metzler2016FromDT,zhao2018cream}, the prior term can be the sparsifying operator corresponding to some pre-defined transform basis, such as discrete cosine transform (DCT) and wavelet \cite{zhao2014image,zhao2016video}. They enjoy the merits of strong convergence and theoretical analysis in most cases but are usually limited in high computational complexity and challenges of tuning parameters \cite{zhao2016nonconvex}.
% but are usually time-consuming due to the high complexity of iterative optimization and face the trouble of choosing optimal computational requirements and parameters \cite{zhao2016nonconvex}.

Recently, fueled by the powerful learning capacity of deep networks, several network-based CS algorithms have been proposed \cite{Kulkarni2016ReconNetNR,sun2020dual}. Apparently compared with model-based methods, they can represent image information flexibly with fast inferences, but the architectures of most network-based methods are empirically designed, and the achievements of traditional algorithms are not fully considered \cite{ren2021adaptive}. More recently, some deep unfolding networks (DUNs) with good interpretability have been proposed to combine network with optimization and train a truncated unfolding inference through an end-to-end learning manner, which has become the mainstream for CS \cite{zhang2018ista,zhang2020optimization,you2021ista,you2021coast,zhang2020amp}.

DUN is usually composed of a fixed number of stages, where each stage corresponds to an iteration. Theoretically, increasing the unfolded iteration number will make the recovered results closer to the original images. However, do we need the whole network trunk to process indiscriminately all images containing different information? In reality, the content of different images is substantially different, and some images are inherently easier to restore than others. Two subjective examples are shown in Fig~\ref{fig:intro}, and we try to analyze whether reducing the module numbers in the stages affects reconstruction performance. The stage numbers are all 25 in different models, and each stage contains a gradient descent module (GDM) and a proximal mapping module (PMM). We do experiments to skip some modules (masked as 0) and execute others (masked as 1) by using selectors on our proposed methods (DP-DUN/DPC-DUN), and count the number of the performed modules (dubbed \emph{active modules}), from which we can see that adaptively selecting a proper path and an optimal number of active modules can maintain similar high performance while largely reducing the computational cost. This observation motivates the possibility of saving computations by selecting a different path for each image based on its content. 
% What is more, the flexibility of the efficiency-accuracy trade-off is needed based on personal preference.

This paper proposes a dynamic path-controllable deep unfolding network (DPC-DUN) for image CS. In DPC-DUN, we design a path-controllable selector (PCS) that comprises a path selector (PS) and a controllable unit (CU), which dynamically adjusts the active module numbers at test time conditioned on the Lagrange multiplier. PS adaptively selects an appropriate path and changes the number and position of the active modules for images with different content. CU takes the Lagrange multiplier as the input and produces a latent representation to control the performance-complexity tradeoff. Our DPC-DUN is a comprehensive framework, considering the merits of previous optimization-inspired DUNs, adjusting the utilization of module numbers, and simplifying the processing pipeline. It enjoys the advantages of both the satisfaction of interpretability and scalability. The major contributions are summarized as follows: 
\begin{itemize}
\item
We propose a novel \textbf{D}ynamic \textbf{P}ath-\textbf{C}ontrollable \textbf{D}eep \textbf{U}nfolding \textbf{N}etwork (\textbf{DPC-DUN}), which shows a slimming mechanism that adaptively selects different paths and enables model interactivity with controllable complexity.
\item
We introduce a \textbf{P}ath \textbf{S}elector (\textbf{PS}) which utilizes the skip connection structure inherent in DUNs and adaptively determines the number and position of modules executed. 
% and facilitates extensibility to other structures.
\item
We design a \textbf{P}ath-\textbf{C}ontrollable \textbf{S}elector (\textbf{PCS}) which achieves dynamic adjustment with a controllable unit (CU) and gets a suitable performance-complexity tradeoff to meet the needs in practice.
% optimize the performance-complexity trade-off of the candidate model architectures.
% \item
% We develop a \textbf{C}ontrollable \textbf{U}nit (\textbf{CU}) that provides the interface for resource/performance modulations with the Lagrange multiplier.
\item
Extensive experiments show that DPC-DUN enables the control of different tradeoffs and achieves good performance.
\end{itemize}

The rest of the paper is organized as follows. Section~\ref{sec:related_work} introduces related work. Then we propose our approach DPC-DUN in Section~\ref{sec3}. Section~\ref{sec4} presents experimental results. Finally, Section~\ref{sec5} draws conclusions.

\section{Related work}
\label{sec:related_work}
\subsection{Deep Unfolding Network}

% \textcolor{magenta}{CS reconstruction methods attract much attention \cite{zymnis2009compressed} and are applied to many tasks \cite{shen2013compressed}. In this paper, we focus on deep unfolding network (DUN) for the CS problem.} 
Deep unfolding networks (DUNs) have been proposed to solve different image inverse tasks, such as denoising \cite{chen2016trainable,lefkimmiatis2017non}, deblurring \cite{kruse2017learning,wang2020stacking}, and demosaicking \cite{kokkinos2018deep}. DUN has friendly interpretability on training data pairs $\{(\mathbf{y}_j, \mathbf{x}_j)\}_{j=1}^{N_a}$, which is usually formulated on CS construction as the bi-level optimization problem:
%%%%%%%%%%%%%%%%%%%%%%%%%%%%%%%%%%%%%%%%%%
% \vspace{-3pt}
\begin{equation}
\begin{split}
&\underset{\Theta}{\min} \sum_{j=1}^{N_a} {\mathcal{L}(\mathbf{\hat{x}}_j, \mathbf{x}_j)}, \\
&\mathbf{s.t.}\ \mathbf{\hat{x}}_j=\underset{\mathbf{x}}{\arg \min } \frac{1}{2} \left \| \mathbf{\Phi}\mathbf{x}-\mathbf{y}_j\right \|^2_2+\lambda\mathcal{R}(\mathbf{x}). 
\end{split}
\end{equation}
%%%%%%%%%%%%%%%%%%%%%%%%%%%%%%%%%%%%%%%%%

%%%%%%%%%%%%%%%%%%%%%%%%%%%%%%%%%%%%%%%%
\begin{figure*}[t]
\centering
\setlength{\abovecaptionskip}{5pt}
\setlength{\belowcaptionskip}{-0.cm}
% \vspace{-0.3cm}
\includegraphics[width=1.0\textwidth]{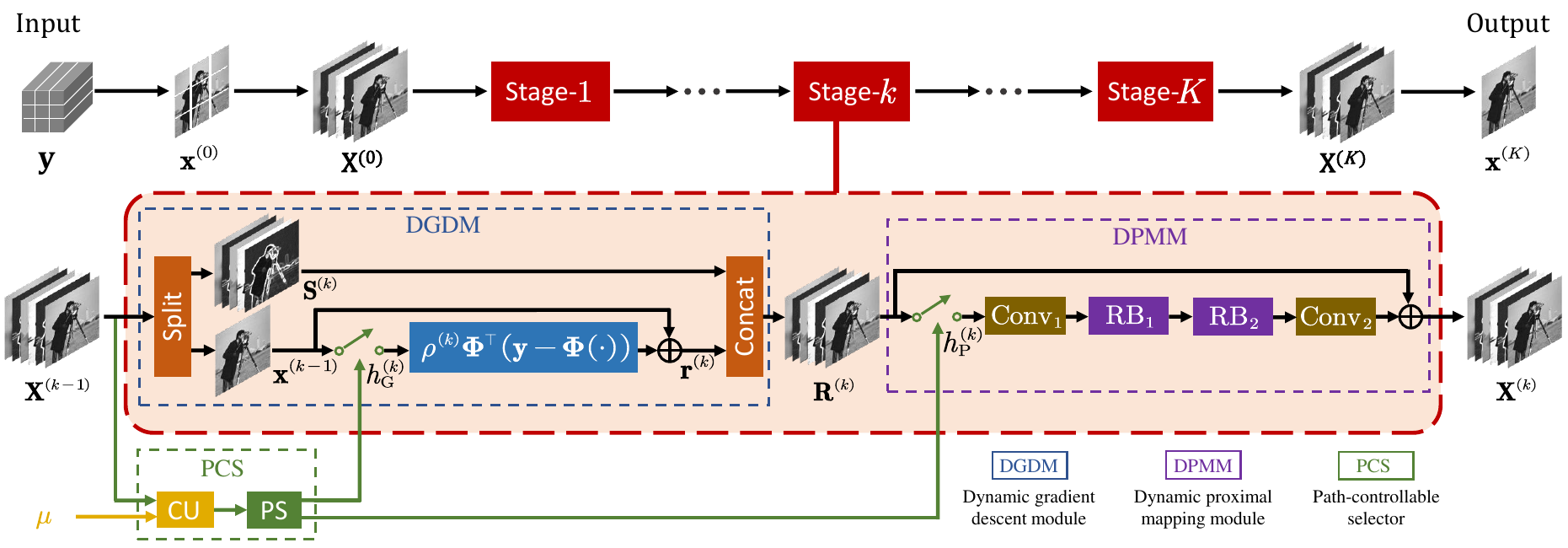} 
\vspace{-0.5cm}
\caption{The architecture of our proposed DPC-DUN which consists of $K$ stages. $\mathbf{y}$ is the under-sampled data as the input of the model, $\mathbf{x}^{(0)}=\mathbf{\Phi^{\top}}\mathbf{y}$ denotes the initialization, $\mathbf{x}^{(K)}$ denotes the recovered result and $\mathbf{X}^{(k)}$ stands for the output features of the $k$-th unrolled stage. In addition to the main processing path marked in black, the green line controls whether the Dynamic Gradient Descent Module (DGDM) and the Dynamic Proximal Mapping Module (DPMM) are selected by the Path-Controllable Selector (PCS) which consists of the Controllable Unit (CU) and the Path Selector (PS), and the yellow line denotes the modulation process with the Lagrange multiplier $\mu$. }
\label{fig:DPC-DUN}
\vspace{-0.5cm}
\end{figure*}
%%%%%%%%%%%%%%%%%%%%%%%%%%%%%%%%%%%%%%%%

DUNs on CS usually integrate some effective convolutional neural network (CNN) denoisers into some optimization methods, including half quadratic splitting (HQS) algorithm \cite{zhang2017learning,dong2018denoising}, proximal gradient descent (PGD) algorithm \cite{zhang2018ista,gilton2019neumann,you2021coast,song2021memory,chenlearning}, and inertial proximal algorithm for nonconvex optimization (iPiano) \cite{su2020ipiano}. Different optimization methods usually lead to different optimization-inspired DUNs. 
% Note that ISTA is much more general than other algorithms, which can work for both general CS and CS-MRI. 
The structures of existing DUNs generally are elaborately designed under a fixed number of iterations and do not consider the effect of different stage numbers in a model, which causes processing of all images with a single path and makes these methods less efficient \cite{yu2021path}.
% smooth images with mild distortions do not need such a deep network to restore and processing all images with a single path makes these methods less efficient \cite{yu2021path}. 
However, the solution has not yet been found to apply in DUNs which are a kind of model based on regularly iterative optimization algorithms. 

\subsection{Dynamic Network}

Dynamic networks \cite{han2021dynamic} have been investigated to achieve a better tradeoff between complexity and performance in different tasks \cite{zhu2021dynamic,li2021dynamic,wu2018blockdrop}. Yu et al. \cite{yu2021path} propose the pathfinder which can dynamically select an appropriate network path to save the computational cost. And some approaches \cite{wang2018skipnet,song2019dynamic} are proposed to skip some residual blocks with skip connection in ResNet \cite{he2016deep} by using a gating network. These methods successfully explore routing policies and touch off the thinking of skip connection. The optimization algorithms applied in DUNs generally have inherent structures of skip connection. Therefore, DUNs have critical physical properties to adaptively achieve the goal of dynamic path selection and minimize the computation burden at the test time.
% Ning et al. \cite{ning2021searching} present an approach to select network width and depth automatically by connecting the design of DUN with NAS, but it does not greatly take advantage of the interpretability. Therefore, it is important to explore the physical properties of DUN to realize an efﬁciency-accuracy tradeoff. 

\subsection{Controllable Network}

Many controllable networks interpolate the parameters to adjust restored effect for various degradation levels \cite{he2021interactive,you2021coast,cai2021toward,jiang2021towards}, which inspire us to design an adjustable model that can be run at different performance-complexity tradeoff on CS reconstruction. Choi et al. \cite{choi2019variable} and Lin et al. \cite{lin2020variable} develop variable-rate learned image compression framework using the control parameters to modulate the internal feature maps in the autoencoder and propose highly flexible models that provide dynamic adjustment of computational cost, thus addressing the main requirements in practical image compression.

\section{Proposed Method}
\label{sec3}

In this section, we will elaborate on the design of our proposed dynamic path-controllable deep unfolding network (DPC-DUN) for image CS.

\subsection{Framework}

As widely recognized knowledge, DUN is generally a typical CNN-based method often constructed with an efficient iterative algorithm. The proximal gradient descent (PGD) algorithm is well-suited for solving many large-scale linear inverse problems. Traditional PGD solves the CS reconstruction problem in Eq.~(\ref{eq: opt}) by iterating between the following two update steps: 
%%%%%%%%%%%%%%%%%%%%%%%%%%%%%%%%%%%%%%%%%%
\begin{align}
\label{eq:r}
&\mathbf{r}^{(k)}=\mathbf{x}^{(k-1)} - \rho^{(k)} \mathbf{\Phi^{\top}} \left(\mathbf{\Phi} \mathbf{x}^{(k-1)} - \mathbf{{y}}\right), \\
\label{eq:x}
&\mathbf{x}^{(k)}=\underset{\mathbf{x}}{\arg \min } \frac{1}{2}\left \|\mathbf{x}-\mathbf{r}^{(k)}\right \|^2_2+\lambda\mathcal{R}(\mathbf{x}),
\end{align}
%%%%%%%%%%%%%%%%%%%%%%%%%%%%%%%%%%%%%%%%%
where $k\in \{1,2,\cdots,K\}$ is the stage index, $\mathbf{x}^{(k)}$ is the output image of the $k$-th stage, $\mathbf{\Phi^{\top}}$ is the transpose of the measurement matrix $\mathbf{\Phi}\in\mathbb{R}^{M \times N}$ and $\rho^{(k)}$ is the learnable descent step size. $\mathbf{y}=\mathbf{\Phi\mathbf{x}}$ is the sampled image produced by an image $\mathbf{x}\in\mathbb{R}^{H \times W}$ which is divided into $\frac{H}{\sqrt{N}} \times \frac{W}{\sqrt{N}}$ non-overlapping image blocks with size of $\sqrt{N} \times \sqrt{N}$.

Eq.~(\ref{eq:r}) is implemented by a gradient descent module (GDM) which is trivial \cite{zhang2018ista} and Eq.~(\ref{eq:x}) is achieved by a proximal mapping module (PMM) which is actually a CNN-based denoiser. Motivated that multi-channel feature transmission well ensures maximum signal flow \cite{song2021memory}, we improve PMM with high-throughput information and the iterative process in the $k$-th stage, as shown in Fig.~\ref{fig:DPC-DUN} (indicated by the black line), can be formulated as:
% is denoted by
%%%%%%%%%%%%%%%%%%%%%%%%%%%%%%%%%%%%%%%%%
\begin{align}
\label{eq:GDM}
&\mathbf{R}^{(k)}=\mathcal{G}_{\operatorname{GDM}}(\mathbf{X}^{(k-1)}), \\
\label{eq:PMM}
&\mathbf{X}^{(k)}=\mathcal{G}_{\operatorname{PMM}}(\mathbf{R}^{(k)}),
\end{align}
%%%%%%%%%%%%%%%%%%%%%%%%%%%%%%%%%%%%%%%%%
where $\mathbf{R}^{(k)}$, $\mathbf{X}^{(k)}$ $\in\mathbb{R}^{C \times H \times W}$ are as the outputs of the $k$-th stage in feature domain. The initialization $\mathbf{X}^{(0)}$ is generated by concatenating the $\mathbf{x}^{(0)}$ and the feature maps ($\in\mathbb{R}^{(C-1) \times H \times W}$) applying a one-convolution layer on the $\mathbf{x}^{(0)}$. 
%%%%%%%%%%%%%%%%%%%%%%%%%%%%%%%%%%%%%%%%
\begin{figure*}[t]
\centering
% \vspace{-0.5cm}
\setlength{\abovecaptionskip}{3pt}
\setlength{\belowcaptionskip}{-0.cm}
\includegraphics[width=1.0\textwidth]{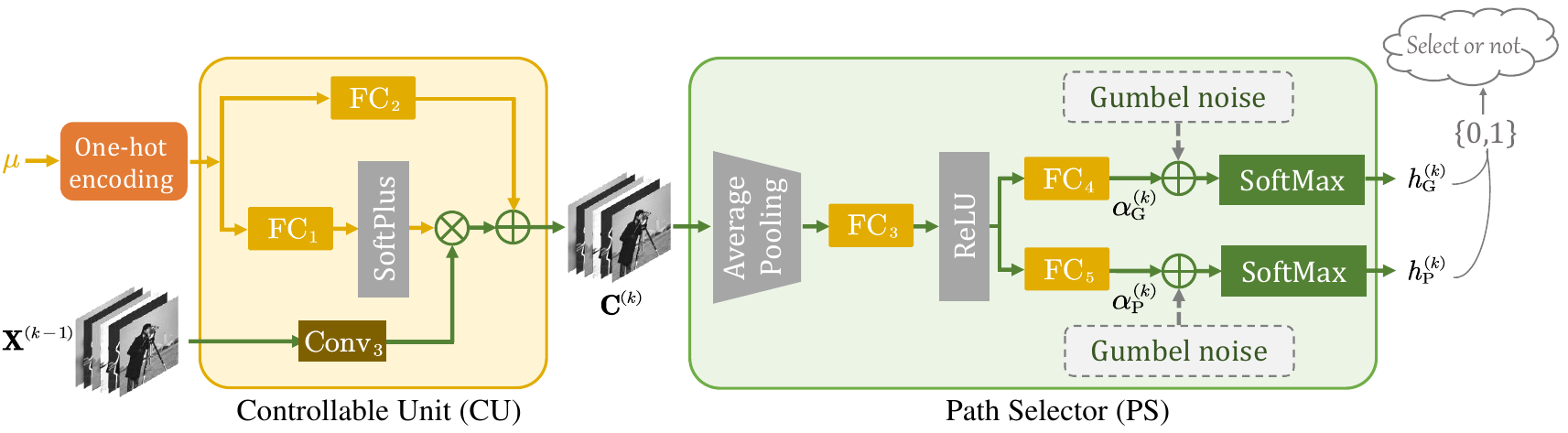} 
\caption{The architecture of our proposed Path-Controllable Selector (PCS) which consists of the controllable unit (CU) and the path selector (PS). The Gumbel noise in the path selector is only added to achieve end-to-end training.}
\label{fig:CPS}
\vspace{-0.5cm}
\end{figure*}
%%%%%%%%%%%%%%%%%%%%%%%%%%%%%%%%%%%%%%%%

To reduce the loss of information, we split the input $\mathbf{X}^{(k-1)}$ into two chunks, dubbed one-channel $\mathbf{x}^{(k-1)}$ which is as the input of Eq.~\eqref{eq:r}, and $(C-1)$-channel $\mathbf{S}^{(k)}$ which is concatenated with the output of Eq.~\eqref{eq:r} along channel dimension. Therefore, the complete process of $\mathcal{G}_{\operatorname{GDM}}(\cdot)$ based on Eq.~\eqref{eq:r} is
%%%%%%%%%%%%%%%%%%%%%%%%%%%%%%%%%%%%%%%%%
\begin{subnumcases}{}
\mathbf{x}^{(k-1)}, \mathbf{S}^{(k)}=\operatorname{Split}\left(\mathbf{X}^{(k-1)}\right), \label{eq:GDM_1} \\
% \mathbf{X}^{(k-1)}[0:1,\ :\ ,\ :\ ], \mathbf{X}^{(k-1)}[1:C,\ :\ ,\ :\ ]\\
\mathbf{r}^{(k)}=\rho^{(k)} \mathbf{\Phi^{\top}} \left(\mathbf{{y}}-\mathbf{\Phi} \mathbf{x}^{(k-1)}\right)+\mathbf{x}^{(k-1)}, \label{eq:GDM_2} \\
\mathbf{R}^{(k)}=\operatorname{Concat}\left(\mathbf{r}^{(k)}, \mathbf{S}^{(k)}\right), \label{eq:GDM_3}
\end{subnumcases}
%%%%%%%%%%%%%%%%%%%%%%%%%%%%%%%%%%%%%%%%%
% %%%%%%%%%%%%%%%%%%%%%%%%%%%%%%%%%%%%%%%%%
% \begin{align}
% \label{eq:GDM_1}
% \mathbf{x}^{(k-1)}, \mathbf{S}^{(k)}&=\operatorname{Split}\left(\mathbf{X}^{(k-1)}\right), \\
% % \mathbf{X}^{(k-1)}[0:1,\ :\ ,\ :\ ], \mathbf{X}^{(k-1)}[1:C,\ :\ ,\ :\ ]\\
% \label{eq:GDM_2}
% \mathbf{r}^{(k)}&=\rho^{(k)} \mathbf{\Phi^{\top}} \left(\mathbf{{y}}-\mathbf{\Phi} \mathbf{x}^{(k-1)}\right)+\mathbf{x}^{(k-1)}, \\
% \label{eq:GDM_3}
% \mathbf{R}^{(k)}&=\operatorname{Concat}\left(\mathbf{r}^{(k)}, \mathbf{S}^{(k)}\right),
% \end{align}
% %%%%%%%%%%%%%%%%%%%%%%%%%%%%%%%%%%%%%%%%%
where $\mathbf{x}^{(k-1)}=\mathbf{X}^{(k-1)}[0:1, :, :\ ], \mathbf{S}^{(k)}=\mathbf{X}^{(k-1)}[1:C, :, :\ ]$. 

Therefore, GDM can be expressed by a formula:
%%%%%%%%%%%%%%%%%%%%%%%%%%%%%%%%%%%%%%%%%
\begin{equation}
% \begin{split}
\label{eq:GDM_total}
\mathbf{R}^{(k)}=\operatorname{Concat}(\rho^{(k)} \mathbf{\Phi^{\top}} \left(\mathbf{{y}}-\mathbf{\Phi}\mathbf{x}^{(k-1)}\right) + \mathbf{x}^{(k-1)}, \ \mathbf{S}^{(k)}).
% \end{split}
\end{equation}
%%%%%%%%%%%%%%%%%%%%%%%%%%%%%%%%%%%%%%%%%

PMM consists of two basic convolution layers ($\operatorname{Conv}_{1}$ and $\operatorname{Conv}_{2}$) and two residual blocks ($\operatorname{RB}_{1}$ and $\operatorname{RB}_{2}$), which keeps the high recovery accuracy with the simple structure \cite{song2021memory}. The module $\mathcal{G}_{\operatorname{PMM}}(\cdot)$ is formulated as:
%%%%%%%%%%%%%%%%%%%%%%%%%%%%%%%%%%%%%%%%%
\begin{align}
\label{eq:PMM_1}
\mathbf{X}^{(k)}=\operatorname{Conv}_{2}(\operatorname{RB}_{2}(\operatorname{RB}_{1}(\operatorname{Conv}_{1}(\mathbf{R}^{(k)}))))+\mathbf{R}^{(k)}.
\end{align}
%%%%%%%%%%%%%%%%%%%%%%%%%%%%%%%%%%%%%%%%%

In DUNs, the stage number $K$ is generally fixed and cannot adaptively adjust based on the input image, which consumes much memory and unnecessary computational cost. Moreover, according to the different needs of users, developing controllable models that can flexibly handle the efﬁciency-accuracy tradeoff is essential and practical. Therefore, we design a \textbf{P}ath-\textbf{C}ontrollable \textbf{S}elector (\textbf{PCS}) applying in GDM and PMM and propose a dynamic gradient descent module (DGDM) and a dynamic proximal mapping module (DPMM).

\subsection{Path-Controllable Selector (PCS)}

As shown in Fig.~\ref{fig:CPS}, the path-controllable selector (PCS) consists of two parts: a controllable unit and a path selector. The  \textbf{P}ath \textbf{S}elector (\textbf{PS}) can dynamically select an optimal path for each image with different content, and the \textbf{C}ontrollable \textbf{U}nit (\textbf{CU}) can enable control of computation and memory while combining with the PS. In the following, we will introduce the PS and the CU, respectively. 
% And to increase the ability of the selectors to modulate network, we also design a special loss function. In the following, we will introduce the PS, the CU and the loss function respectively. 

\subsubsection{Path Selector (PS)}

The motivation of the path selector (PS) aims at predicting a one-hot vector, which denotes whether to execute or skip the module (\textit{i.e.}, the decision is to skip the module when PS predicts 0.). Since a non-differentiable problem exists in the process from the continuous feature outputs to discrete path selection, we adopt the Gumbel Softmax trick \cite{jang2016categorical,zhu2021dynamic,dai2021mix} to make the discrete decision differentiable during the back-propagation. 

To meet our needs, we first choose two resolution candidates $\alpha^{(k)}_{1}$ and $\alpha^{(k)}_{2}$ at the $k$-th stage to shrink the exploration range. $\alpha^{(k)}=[\alpha^{(k)}_{1},\alpha^{(k)}_{2}]\in\mathbb{R}^2$ is the output signal that represents the probability of each candidate. Then, we apply Gumbel Softmax trick to turn soft decisions $\alpha^{(k)}$ into hard decisions $h^{(k)}=[h^{(k)}_{1},h^{(k)}_{2}]$, and the expression can be written as:
%%%%%%%%%%%%%%%%%%%%%%%%%%%%%%%%%%%%%%%%%
\begin{equation}
% \begin{split}
\label{eq:gumbel_softmax}
h^{(k)}_{i} = \frac{\exp{(\alpha^{(k)}_{i}+n^{(k)}_{i}/\tau)}}{\sum_{i=1}^{2}\exp{(\alpha^{(k)}_{i}+n^{(k)}_{i}/\tau)}}, 
% \mathbf{X}^{(k)}=\operatorname{Conv}_{2}(\operatorname{RB}_{2}(\operatorname{RB}_{1}(\operatorname{Conv}_{1}(\mathbf{R}^{(k)}))))+\mathbf{R}^{(k)}.
% \end{split}
\end{equation}
%%%%%%%%%%%%%%%%%%%%%%%%%%%%%%%%%%%%%%%%%
% For the selection signal $\boldsymbol{\alpha}^{(k)}$ which determines the sampling probability of module at the $k$-th stage, the one-hot selector $\boldsymbol{h}^{(k)}$ is obtained by Gumbel Softmax function as follows:
% %%%%%%%%%%%%%%%%%%%%%%%%%%%%%%%%%%%%%%%%%
% \begin{equation}
% % \begin{split}
% \label{eq:gumbel_softmax}
% h^{(k)}_{i} = \operatorname{GumbelSoftmax}(\alpha^{(k)}_{i}|\boldsymbol{\alpha}^{(k)}) = \frac{\exp{(\alpha^{(k)}_{i}+n^{(k)}_{i}/\tau)}}{\sum_{i}\exp{(\alpha^{(k)}_{i}+n^{(k)}_{i}/\tau)}}, 
% % \mathbf{X}^{(k)}=\operatorname{Conv}_{2}(\operatorname{RB}_{2}(\operatorname{RB}_{1}(\operatorname{Conv}_{1}(\mathbf{R}^{(k)}))))+\mathbf{R}^{(k)}.
% % \end{split}
% \end{equation}
% %%%%%%%%%%%%%%%%%%%%%%%%%%%%%%%%%%%%%%%%%
where $h^{(k)}_{i} \in \{0,1 \}$. $n^{(k)}_{i} \sim \operatorname{Gumbel}(0,1)$ is a random noise sampled from the Gumbel distribution, which is only adopted in the training phase. $\tau$ is a temperature parameter to inﬂuence the Gumbel Softmax function and set it as 1. 

Inspired by squeeze-and-excitation (SE) network \cite{hu2018squeeze}, we design two \textbf{PSs} (${h}^{(k)}_{\text{G}}$ and ${h}^{(k)}_{\text{P}}$) with the same structure to control GDM and PMM respectively, as shown in Fig.~\ref{fig:CPS}. The selectors ${h}^{(k)}_{\text{G}}=\mathcal{H}_{\operatorname{G}}(\mathbf{C^{(k)}})$ and ${h}^{(k)}_{\text{P}}=\mathcal{H}_{\operatorname{P}}(\mathbf{C^{(k)}})$ are formulated as:
%%%%%%%%%%%%%%%%%%%%%%%%%%%%%%%%%%%%%%%%%
\begin{align}
\label{eq:H_g}
{h}^{(k)}_{\text{G}} = \operatorname{GS}(\operatorname{FC}^{(k)}_{4}(\operatorname{ReLU}(\operatorname{FC}^{(k)}_{3}(\operatorname{AveragePool}(\mathbf{C}^{(k)}))))), \\
\label{eq:H_p}
{h}^{(k)}_{\text{P}} = \operatorname{GS}(\operatorname{FC}^{(k)}_{5}(\operatorname{ReLU}(\operatorname{FC}^{(k)}_{3}(\operatorname{AveragePool}(\mathbf{C}^{(k)}))))).
\end{align}
%%%%%%%%%%%%%%%%%%%%%%%%%%%%%%%%%%%%%%%%%
The input feature $\mathbf{C^{(k)}}$ is firstly squeezed by an average pooling operation in each channel to produce the $C$-channel feature maps. Then we reduce the feature dimensions by 4 in the fully-connected layer $\operatorname{FC^{(k)}_3}(\cdot)$. The non-linear activation function ($\operatorname{ReLU}(\cdot)$) and the fully-connected layer ($\operatorname{FC^{(k)}_4}(\cdot)$ or $\operatorname{FC^{(k)}_5}(\cdot)$) are further leveraged to generate the resolution candidate $\alpha^{(k)}_{\text{G}}$ or $\alpha^{(k)}_{\text{P}}$. Finally, the one-hot selections ${h}^{(k)}_{\text{G}}=[{h}^{(k)}_{\text{G1}},{h}^{(k)}_{\text{G2}}] \in \{ 0,1 \}^{2}, \ {h}^{(k)}_{\text{P}}=[{h}^{(k)}_{\text{P1}}, {h}^{(k)}_{\text{P2}}] \in \{ 0,1 \}^{2}$ are obtained by the Gumbel Softmax function $(\operatorname{GS}(\cdot))$.

Eq.~\eqref{eq:GDM_2} and Eq.~\eqref{eq:PMM_1} that both use the residual connection are the primary sources of computation requirement in the main process. Inspired by the controllable residual connection in CResMD \cite{he2021interactive}, we add the two PSs to control the summation weight, respectively. Therefore, the network adding the dynamical PSs, dubbed \textbf{DP-DUN}, is iterated by following two update steps:

%%%%%%%%%%%%%%%%%%%%%%%%%%%%%%%%%%%%%%%%%
\vspace{-10pt}
\begin{small}
\begin{equation}
\begin{split}
\label{eq:dpdun_r}
\mathbf{R}^{(k)}=&\operatorname{Concat}(h^{(k)}_{\text{G1}} \cdot \rho^{(k)} \mathbf{\Phi^{\top}} \left(\mathbf{{y}}-\mathbf{\Phi} \mathbf{x}^{(k-1)}\right)+h^{(k)}_{\text{G2}} \cdot \mathbf{x}^{(k-1)}, \mathbf{S}^{(k)}), \\
% \label{eq:dpdun_x}
\mathbf{X}^{(k)}=&\ h^{(k)}_{\text{P1}} \cdot \operatorname{Conv}_{2}(\operatorname{RB}_{2}(\operatorname{RB}_{1}(\operatorname{Conv}_{1}(\mathbf{R}^{(k)}))))+h^{(k)}_{\text{P2}} \cdot \mathbf{R}^{(k)}.    
\end{split}
\end{equation}
\end{small}
%%%%%%%%%%%%%%%%%%%%%%%%%%%%%%%%%%%%%%%%%
% %%%%%%%%%%%%%%%%%%%%%%%%%%%%%%%%%%%%%%%%%
% \vspace{-10pt}
% \begin{small}
% \begin{equation}
% \begin{split}
% \label{eq:dpdun_r}
% \mathbf{R}^{(k)}=&\operatorname{Concat}(h^{(k)}_{\text{G}} \cdot \rho^{(k)} \mathbf{\Phi^{\top}} \left(\mathbf{{y}}-\mathbf{\Phi} \mathbf{x}^{(k-1)}\right)+\mathbf{x}^{(k-1)}, \mathbf{S}^{(k)}), \\
% % \label{eq:dpdun_x}
% \mathbf{X}^{(k)}=&\ h^{(k)}_{\text{P}} \cdot \operatorname{Conv}_{2}(\operatorname{RB}_{2}(\operatorname{RB}_{1}(\operatorname{Conv}_{1}(\mathbf{R}^{(k)}))))+\mathbf{R}^{(k)}.    
% \end{split}
% \end{equation}
% \end{small}
% %%%%%%%%%%%%%%%%%%%%%%%%%%%%%%%%%%%%%%%%%
Thus, the residual parts can be skipped by setting $\{{h}^{(k)}_{\text{G1}}=0, \ {h}^{(k)}_{\text{G2}}=1\}$ or $\{{h}^{(k)}_{\text{P1}}=0, \ {h}^{(k)}_{\text{P2}}=1\}$ and instead, they are performed. 

In real-world scenarios, physical devices often impose different computing resource constraints on the model. Thus, the network should be developed considering various computational expenses. However, if we use the reconstruction loss only, the selector will tend to provide a sub-optimal solution \cite{zhu2021dynamic}, which conducts as many GDMs and PMMs as possible because features with the most enriched spatial information correspond to relatively higher reconstruction accuracy. To achieve a satisfactory efﬁciency-accuracy tradeoff, we propose a new selection loss $\mathcal{L}_{select}$ to guide the training. The loss $\mathcal{L}_{select}$ is designed by calculating the mean of the two PSs to obtain the selector prediction as:
%%%%%%%%%%%%%%%%%%%%%%%%%%%%%%%%%%%%%%%%%
% \vspace{-10pt}
\begin{equation}
\small
\label{eq:loss_select}
\mathcal{L}_{select} =\ \frac{1}{K}\sum_{k=1}^{K} (h^{(k)}_{\text{G}}+h^{(k)}_{\text{P}}), 
\end{equation}
%%%%%%%%%%%%%%%%%%%%%%%%%%%%%%%%%%%%%%%%%%
where $K$ represents the stage number. Therefore, the total loss function based on the reconstruction loss function $\mathcal{L}_{rec}$ can be optimized as follows:
%%%%%%%%%%%%%%%%%%%%%%%%%%%%%%%%%%%%%%%%%%
\begin{equation} 
\label{eq: loss_total}
\mathcal{L} = \mathcal{L}_{rec} + \mu \mathcal{L}_{select}, 
\end{equation}
%%%%%%%%%%%%%%%%%%%%%%%%%%%%%%%%%%%%%%%%
where the scalar factor $\mu > 0$ in the Lagrangian is called a Lagrange multiplier to balance the accuracy expectation and computational cost constraints. As seen from the experimental results in Section~\ref{sec:mu} , the value of $\mu$ decides different efficiency-accuracy tradeoffs, which is challenging to address the requirements in practice precisely. Therefore, we design a controllable unit (CU) to control the PS and propose DPC-DUN, which alternates between training the model and adjusting the different $\mu$ by introducing $\mu$.

\subsubsection{Controllable Unit (CU)}

% For achieving an $\lambda$-scheduling algorithm, controllable path selector (CPS) introduces a controllable convolution based on PS, in Fig.~\ref{fig:CPS}. 
In Fig.~\ref{fig:CPS}, the Lagrange multiplier $\mu$ is firstly encoded as one-hot vector and then inputs two functions to get a parameter pair $(\mathbf{Q}^{(k)}(\mu), \mathbf{P}^{(k)}(\mu))$, which can be expressed as:
%%%%%%%%%%%%%%%%%%%%%%%%%%%%%%%%%%%%%%%%%%
\begin{align}
% \small
% \setlength{\abovedisplayskip}{10pt}
% \setlength{\belowdisplayskip}{5pt}
&\mathbf{Q}^{(k)}(\mu) = \operatorname{Softplus}(\operatorname{FC}_{1}(\operatorname{onehot}(\mu))), \\
&\mathbf{P}^{(k)}(\mu) = \operatorname{FC}_{2}(\operatorname{onehot}(\mu)),
% \vspace{-3pt}
\end{align}
%%%%%%%%%%%%%%%%%%%%%%%%%%%%%%%%%%%%%%%%%
where $\operatorname{FC}_{1}(\cdot)$ and $\operatorname{FC}_{2}(\cdot)$ are the fully-connected layers, $\operatorname{Softplus}(x)=\log(1+\exp(x))$. After obtaining the pair, we use the affine transformation to scale and shift the element-wise feature maps, which is shown as:
%%%%%%%%%%%%%%%%%%%%%%%%%%%%%%%%%%%%%%%%%%
\begin{align}
% \small
% \setlength{\abovedisplayskip}{10pt}
% \setlength{\belowdisplayskip}{5pt}
\mathbf{C}^{(k)} = \mathbf{Q}^{(k)}(\mu) \otimes \operatorname{Conv}_{3}(\mathbf{X}^{(k-1)}) + \mathbf{P}^{(k)}(\mu),
% \vspace{-3pt}
\end{align}
%%%%%%%%%%%%%%%%%%%%%%%%%%%%%%%%%%%%%%%%%
The controllable unit, denoted by $\mathbf{C}^{(k)}=\mathcal{H}_{\operatorname{CU}}(\mathbf{X}^{(k-1)}, \mu)$, is expressed as:
%%%%%%%%%%%%%%%%%%%%%%%%%%%%%%%%%%%%%%%%%%
\begin{small}
\begin{equation}
\begin{split}
% \small
% \setlength{\abovedisplayskip}{10pt}
% \setlength{\belowdisplayskip}{5pt}
\mathbf{C}^{(k)} =& \operatorname{Softplus}(\operatorname{FC}_{1}(\operatorname{onehot}(\mu))) \otimes \operatorname{Conv}_{3}(\mathbf{X}^{(k-1)}) \\
&+ \operatorname{FC}_{2}(\operatorname{onehot}(\mu)).
% \vspace{-3pt}
\end{split}
\end{equation}
\end{small}
%%%%%%%%%%%%%%%%%%%%%%%%%%%%%%%%%%%%%%%%%
Hence, our proposed path-controllable selectors (\textbf{PCSs}) consisting of the controllable unit (CU) and the path selector (PS) can be formulated as:
%%%%%%%%%%%%%%%%%%%%%%%%%%%%%%%%%%%%%%%%%
\begin{equation} 
\begin{split}
\label{eq:cps_g}
&{h}^{(k)}_{\text{G}}(\mu) = \mathcal{H}_{\operatorname{G}}(\mathcal{H}_{\operatorname{CU}}(\mathbf{X}^{(k-1)}, \mu)), \\
% \label{eq:cps_p}
&{h}^{(k)}_{\text{P}}(\mu) = \mathcal{H}_{\operatorname{P}}(\mathcal{H}_{\operatorname{CU}}(\mathbf{X}^{(k-1)}, \mu)).
\end{split}
\end{equation} 
%%%%%%%%%%%%%%%%%%%%%%%%%%%%%%%%%%%%%%%%%

Our proposed \textbf{D}ynamic \textbf{P}ath-\textbf{C}ontrollable \textbf{D}eep \textbf{U}nfolding \textbf{N}etwork (\textbf{DPC-DUN}) solves the CS reconstruction problem by iterating the dynamic gradient descent module (DGDM) and the dynamic proximal mapping module (DPMM) which are both adding PCS based on GDM and PMM respectively. According, $\mathbf{R}^{(k)}=\mathcal{G}_{\operatorname{DGDM}}(\mathbf{X}^{(k-1)}, \mu)$ and $\mathbf{X}^{(k)}=\mathcal{G}_{\operatorname{DPMM}}(\mathbf{R}^{(k)}, \mu)$ can be expressed as:

%%%%%%%%%%%%%%%%%%%%%%%%%%%%%%%%%%%%%%%%%
\vspace{-10pt}
\begin{small}
\begin{equation}
\begin{split}
\label{eq:dpcdun}
\mathbf{R}^{(k)}=&\operatorname{Concat}(h^{(k)}_{\text{G1}}(\mu) \cdot \rho^{(k)} \mathbf{\Phi^{\top}} \left(\mathbf{{y}}-\mathbf{\Phi} \mathbf{x}^{(k-1)}\right)+h^{(k)}_{\text{G2}}(\mu) \cdot \mathbf{x}^{(k-1)}, 
\\ &\mathbf{S}^{(k)}),   
% \label{eq:dpcdun_x}
\\  \mathbf{X}^{(k)}=&\ h^{(k)}_{\text{P1}}(\mu) \cdot \operatorname{Conv}_{2}(\operatorname{RB}_{2}(\operatorname{RB}_{1}(\operatorname{Conv}_{1}(\mathbf{R}^{(k)}))))+h^{(k)}_{\text{P2}}(\mu) \cdot \mathbf{R}^{(k)}.
\end{split}
\end{equation}
\end{small}
%%%%%%%%%%%%%%%%%%%%%%%%%%%%%%%%%%%%%%%%%

\subsection{Loss function}

The reconstruction loss $\mathcal{L}_{rec}$ in this work is the $L_1$ loss function, denoted by
% %%%%%%%%%%%%%%%%%%%%%%%%%%%%%%%%%%%%%%%%%%
% \begin{equation} 
% % \setlength{\abovedisplayskip}{5pt}
% % \setlength{\belowdisplayskip}{6pt}
% \label{eq: loss_rec}
% \mathcal{L}_{rec} =\ \frac{1}{{N}{N_a}}\sum_{j=1}^{N_a} {\left\|\mathbf{x}_j-\mathbf{x}_j^{(K)}\right\|}_1, 
% \end{equation}
% %%%%%%%%%%%%%%%%%%%%%%%%%%%%%%%%%%%%%%%%
%%%%%%%%%%%%%%%%%%%%%%%%%%%%%%%%%%%%%%%%%%
\begin{equation} 
\label{eq: loss_rec}
\mathcal{L}_{rec}(\mu) =\ \frac{1}{{N}{N_a}}\sum_{j=1}^{N_a} {\left\|\mathbf{x}_j-\mathbf{x}_j^{(K)}(\mu)\right\|}_1, 
\end{equation}
%%%%%%%%%%%%%%%%%%%%%%%%%%%%%%%%%%%%%%%%
where $N_a$ represents the number of training images, $N$ represents the size of each image, $\left\{\mathbf{x}_j\right\}_{j=1}^{N_a}$ is a set of full-sampled images and $\left\{\mathbf{x}_j^{(K)}(\mu)\right\}_{j=1}^{N_a}$ is the corresponding reconstruction result with the controlling coefficient $\mu$.

Based on Eq.~\eqref{eq:loss_select}, the selection loss function is updated to be
%%%%%%%%%%%%%%%%%%%%%%%%%%%%%%%%%%%%%%%%%%
\begin{equation} 
\label{eq: loss_select_mu}
\mathcal{L}_{select}(\mu) =\ \frac{1}{K}\sum_{k=1}^{K} (h^{(k)}_{\text{G}}(\mu)+h^{(k)}_{\text{P}}(\mu)).
\end{equation}
%%%%%%%%%%%%%%%%%%%%%%%%%%%%%%%%%%%%%%%%
% Therefore, the selection loss and the reconstruction loss adding the Lagrange multiplier $\mu$ are denoted by
% %%%%%%%%%%%%%%%%%%%%%%%%%%%%%%%%%%%%%%%%%
% \begin{align}
% % \label{eq:loss_select}
% \mathcal{L}_{select}(\mu) =\ \frac{1}{K}\sum_{k=1}^{K} (h^{(k)}_{\text{G}}(\mu)+h^{(k)}_{\text{P}}(\mu)), \\
% % \label{eq:loss_select}
% \mathcal{L}_{rec}(\mu) =\ \frac{1}{{N}{N_a}}\sum_{j=1}^{N_a} {\left\|\mathbf{x}_j-\mathbf{x}_j^{(K)}(\mu)\right\|}_1.
% \end{align}
% %%%%%%%%%%%%%%%%%%%%%%%%%%%%%%%%%%%%%%%%%

Therefore, the total loss function can be expressed as below:
%%%%%%%%%%%%%%%%%%%%%%%%%%%%%%%%%%%%%%%%%%
\begin{equation} 
% \setlength{\abovedisplayskip}{5pt}
% \setlength{\belowdisplayskip}{6pt}
% \label{eq: loss_total_dpc}
\mathcal{L}(\mu) = \mathcal{L}_{rec}(\mu) + \mu \mathcal{L}_{select}(\mu). 
\end{equation}
%%%%%%%%%%%%%%%%%%%%%%%%%%%%%%%%%%%%%%%%

\section{Experiments}\label{sec4}

\subsection{Implementation Details}

%%%%%%%%%%%%%%%%%%%%%%%%%%%%%%%%%%%%
\begin{table*}[t]
		\centering
		\small
		\caption{Average PSNR(dB)/SSIM performance comparisons on Set11 dataset \cite{Kulkarni2016ReconNetNR} with different CS ratios. We compare our methods with nine competing methods. The best and second best PSNR/SSIM results are highlighted in \textcolor{red}{red} and \textcolor{blue}{blue} colors, respectively. The number of GDM/PMM executed is highlighted in \textcolor{cyan}{cyan}.}
		\label{tab:set11}
		\vspace{-5pt}
		\resizebox{\textwidth}{!}{%
		\renewcommand\tabcolsep{8pt}
			\begin{tabular}{c|cc|cc|cc|cc|cc}
			\toprule
				% \hline
				\multicolumn{1}{c|}{\multirow{2}{*}{CS Ratio}}&\multicolumn{2}{c}{10\%}&\multicolumn{2}{c}{25\%}&\multicolumn{2}{c}{30\%}&\multicolumn{2}{c}{40\%}&\multicolumn{2}{c}{50\%}\\ \cline{2-11}
                \multicolumn{1}{c|}{}&PSNR&SSIM&PSNR&SSIM&PSNR&SSIM&PSNR&SSIM&PSNR&SSIM \\
				% \cline{3-7}
				% &&10\%&25\%&30\%&40\%&50\%\\
				\toprule
				% \hline
				% \hline
				% \multirow{10}{*}{Set11}
				\multicolumn{1}{l|}{IRCNN \cite{zhang2017learning}} &23.05&0.6789&28.42&0.8382&29.55&0.8606&31.30&0.8898&32.59&0.9075\\
				\multicolumn{1}{l|}{ReconNet \cite{Kulkarni2016ReconNetNR}} &23.96&0.7172&26.38&0.7883&28.20&0.8424&30.02&0.8837&30.62&0.8983\\
				\multicolumn{1}{l|}{GDN \cite{gilton2019neumann}} &23.90&0.6927&29.20&0.8600&30.26&0.8833&32.31&0.9137&33.31&0.9285\\
				\multicolumn{1}{l|}{DPDNN \cite{dong2018denoising}} &26.23&0.7992&31.71&0.9153&33.16&0.9338&35.29&0.9534&37.63&0.9693\\
				\multicolumn{1}{l|}{ISTA-Net$^+$ \cite{zhang2018ista}} &26.58&0.8066&32.48&0.9242&33.81&0.9393&36.04&0.9581&38.06&0.9706\\
				\multicolumn{1}{l|}{DPA-Net \cite{sun2020dual}} &27.66&0.8530&32.38&0.9311&33.35&0.9425&35.21&0.9580&36.80&0.9685\\ 
				\multicolumn{1}{l|}{MAC-Net \cite{chenlearning}} &27.68&0.8182&32.91&0.9244&33.96&0.9372&35.94&0.9560&37.67&0.9668\\
				\multicolumn{1}{l|}{iPiano-Net \cite{su2020ipiano}} &28.33&0.8548&33.60&0.9368&34.90&0.9487&37.02&0.9638&38.85&0.9736\\
				\multicolumn{1}{l|}{COAST \cite{you2021coast}} &28.74&0.8619&33.98&0.9407&35.11&0.9505&37.11&0.9646&38.94&0.9744\\
                % \multicolumn{1}{l|}{HQSRED-Net \cite{ma2022deep}}&29.04&0.8678&-&-&35.58&0.9553&37.68&0.9683&\textcolor{blue}{39.89}&\textcolor{blue}{0.9784}\\
                % \multicolumn{1}{l|}{SCS-GNet \cite{zhong2022scalable}}&29.35&\textcolor{red}{0.8854}&-&-&35.42&\textcolor{red}{0.9588}&37.55&\textcolor{red}{0.9705}&39.29&0.9769\\
                \cline{1-11}
				
				\multicolumn{1}{l|}{\multirow{2}{*}{DP-DUN}}
				&\textcolor{red}{29.42}&\textcolor{red}{0.8806}&\textcolor{red}{34.75}&\textcolor{red}{0.9483}&\textcolor{red}{36.02}&\textcolor{red}{0.9577}&\textcolor{red}{38.06}&\textcolor{red}{0.9698}&\textcolor{red}{39.90}&\textcolor{red}{0.9791}\\
				\multicolumn{1}{c|}{}&\multicolumn{2}{c|}{\textcolor{cyan}{(22.9/23.2)}}&\multicolumn{2}{c|}{\textcolor{cyan}{(19.0/20.7)}}&\multicolumn{2}{c|}{\textcolor{cyan}{(21.0/22.9)}}&\multicolumn{2}{c|}{\textcolor{cyan}{(19.0/21.4)}}&\multicolumn{2}{c}{\textcolor{cyan}{(17.0/20.9)}}\\ 
				
				\multicolumn{1}{l|}{\multirow{2}{*}{DPC-DUN}}
				&\textcolor{blue}{29.40}&\textcolor{blue}{0.8798}&\textcolor{blue}{34.69}&\textcolor{blue}{0.9482}&\textcolor{blue}{35.88}&\textcolor{blue}{0.9570}&\textcolor{blue}{37.98}&\textcolor{blue}{0.9694}&\textcolor{blue}{39.84}&\textcolor{blue}{0.9778}\\
				\multicolumn{1}{c|}{}&\multicolumn{2}{c|}{\textcolor{cyan}{(20.5/21.6)}}&\multicolumn{2}{c|}{\textcolor{cyan}{(17.8/18.9)}}&\multicolumn{2}{c|}{\textcolor{cyan}{(16.5/18.3)}}&\multicolumn{2}{c|}{\textcolor{cyan}{(16.5/18.6)}}&\multicolumn{2}{c}{\textcolor{cyan}{(14.6/17.5)}}\\
				\toprule
				% \hline
				% \hline
		\end{tabular}}
		\vspace{-0.3cm}
	\end{table*}
%%%%%%%%%%%%%%%%%%%%%%%%%%%%%%%%%%%%
%%%%%%%%%%%%%%%%%%%%%%%%%%%%%%%%%%%%
\begin{table*}[t]
		\centering
		\small
		\caption{Average PSNR(dB)/SSIM performance comparisons on CBSD68 dataset \cite{zhang2021plug} with different CS ratios. We compare our methods with nine competing methods. The best and second best PSNR/SSIM results are highlighted in \textcolor{red}{red} and \textcolor{blue}{blue} colors, respectively. The number of GDM/PMM executed is highlighted in \textcolor{cyan}{cyan}.}
		\label{tab:cbsd68}
		\vspace{-5pt}
		\resizebox{\textwidth}{!}{%
		\renewcommand\tabcolsep{8pt}
			\begin{tabular}{c|cc|cc|cc|cc|cc}
			\toprule
				% \hline
				\multicolumn{1}{c|}{\multirow{2}{*}{CS Ratio}}&\multicolumn{2}{c}{10\%}&\multicolumn{2}{c}{25\%}&\multicolumn{2}{c}{30\%}&\multicolumn{2}{c}{40\%}&\multicolumn{2}{c}{50\%}\\ \cline{2-11}
                \multicolumn{1}{c|}{}&PSNR&SSIM&PSNR&SSIM&PSNR&SSIM&PSNR&SSIM&PSNR&SSIM \\
				% \cline{3-7}
				% &&10\%&25\%&30\%&40\%&50\%\\
				\toprule
				% \hline
				% \hline
				% \multirow{10}{*}{Set11}
				\multicolumn{1}{l|}{IRCNN \cite{zhang2017learning}} &23.07&0.5580&26.44&0.7206&27.31&0.7543&28.76&0.8042&30.00&0.8398\\
				\multicolumn{1}{l|}{ReconNet \cite{Kulkarni2016ReconNetNR}} &24.02&0.6414&26.01&0.7498&27.20&0.7909&28.71&0.8409&29.32&0.8642\\
				\multicolumn{1}{l|}{GDN \cite{gilton2019neumann}} &23.41&0.6011&27.11&0.7636&27.52&0.7745&30.14&0.8649&30.88&0.8763\\
				\multicolumn{1}{l|}{DPDNN \cite{dong2018denoising}} &25.35&0.7020&29.28&0.8513&30.39&0.8807&32.21&0.9171&34.27&0.9455\\
				\multicolumn{1}{l|}{ISTA-Net$^+$ \cite{zhang2018ista}} &25.37&0.7022&29.32&0.8515&30.37&0.8786&32.23&0.9165&34.04&0.9425\\
				\multicolumn{1}{l|}{DPA-Net \cite{sun2020dual}} &25.47&0.7365&29.01&0.8589&29.73&0.8821&31.17&0.9151&32.55&0.9382\\ 
				\multicolumn{1}{l|}{MAC-Net \cite{chenlearning}} &25.80&0.7018&29.42&0.8464&30.28&0.8707&32.02&0.9074&33.68&0.9348\\
				\multicolumn{1}{l|}{iPiano-Net \cite{su2020ipiano}} &26.34&0.7431&30.16&0.8711&31.24&0.8964&33.14&0.9298&34.98&0.9521\\
				\multicolumn{1}{l|}{COAST \cite{you2021coast}} &26.43&0.7450&30.24&0.8724&31.20&0.8947&33.03&0.9273&34.81&0.9497\\ \cline{1-11}
				
				\multicolumn{1}{l|}{\multirow{2}{*}{DP-DUN}}
				&\textcolor{red}{26.82}&\textcolor{blue}{0.7601}&\textcolor{red}{30.76}&\textcolor{red}{0.8837}&\textcolor{red}{31.83}&\textcolor{red}{0.9059}&\textcolor{red}{33.77}&\textcolor{red}{0.9369}&\textcolor{red}{35.72}&\textcolor{red}{0.9581}\\
				\multicolumn{1}{c|}{}&\multicolumn{2}{c|}{\textcolor{cyan}{(23.0/23.4)}}&\multicolumn{2}{c|}{\textcolor{cyan}{(19.0/21.0)}}&\multicolumn{2}{c|}{\textcolor{cyan}{(21.0/22.7)}}&\multicolumn{2}{c|}{\textcolor{cyan}{(18.8/21.1)}}&\multicolumn{2}{c}{\textcolor{cyan}{(17.0/20.2)}}\\ 
				
				\multicolumn{1}{l|}{\multirow{2}{*}{DPC-DUN}}
				&\textcolor{blue}{26.79}&\textcolor{red}{0.7611}&\textcolor{blue}{30.71}&\textcolor{blue}{0.8828}&\textcolor{blue}{31.76}&\textcolor{blue}{0.9051}&\textcolor{blue}{33.70}&\textcolor{blue}{0.9364}&\textcolor{blue}{35.62}&\textcolor{blue}{0.9573}\\
				\multicolumn{1}{c|}{}&\multicolumn{2}{c|}{\textcolor{cyan}{(19.6/21.6)}}&\multicolumn{2}{c|}{\textcolor{cyan}{(17.7/19.4)}}&\multicolumn{2}{c|}{\textcolor{cyan}{(17.2/18.5)}}&\multicolumn{2}{c|}{\textcolor{cyan}{(16.2/18.0)}}&\multicolumn{2}{c}{\textcolor{cyan}{(14.3/16.9)}}\\
				\toprule
				% \hline
				% \hline
		\end{tabular}}
		\vspace{-0.3cm}
	\end{table*}
%%%%%%%%%%%%%%%%%%%%%%%%%%%%%%%%%%%%
%%%%%%%%%%%%%%%%%%%%%%%%%%%%%%%%%%%%
\begin{table*}[t]
		\centering
		\setlength{\abovecaptionskip}{0pt}
        \setlength{\belowcaptionskip}{1pt}
		% \small
		\caption{Average PSNR(dB)/SSIM performance comparisons on Urban100 \cite{dong2018denoising} and DIV2K \cite{Agustsson_2017_CVPR_Workshops} datasets with different CS ratios. We compare our methods with three competing methods. The best and second best results are highlighted in \textcolor{red}{red} and \textcolor{blue}{blue} colors, respectively. The numbers of GDM/PMM executed are highlighted in \textcolor{cyan}{cyan}.}
		\label{tab:urban100_div2k}
% 		\vspace{-5pt}
		\resizebox{\textwidth}{!}{%
			\begin{tabular}{c|c|cccccc}
			\toprule
				% \hline
				\multicolumn{1}{c|}{\multirow{2}{*}{Dataset}}&\multicolumn{1}{c|}{\multirow{2}{*}{Methods}}&\multicolumn{6}{c}{CS Ratio}\\ 
				\cline{3-8}
				\multicolumn{1}{c|}{}&\multicolumn{1}{c|}{}&\multicolumn{1}{c}{5\%}  &\multicolumn{1}{c}{10\%}&\multicolumn{1}{c}{25\%}&\multicolumn{1}{c}{30\%}&\multicolumn{1}{c}{40\%}&\multicolumn{1}{c}{50\%}\\ 
				% \cline{3-7}
				% &&10\%&25\%&30\%&40\%&50\%\\
				\toprule
				% \hline
				% \hline
				\multirow{7}{*}{Urban100}
				&\multicolumn{1}{l|}{ISTA-Net$^+$ \cite{zhang2018ista}} &20.70/0.5886&23.61/0.7238&28.93/0.8840&30.21/0.9079&32.43/0.9377&34.43/0.9571\\
				&\multicolumn{1}{l|}{iPiano-Net \cite{su2020ipiano}} &22.48/0.6728&25.67/0.7963&30.87/0.9157&32.16/0.9320&34.27/0.9531&36.22/0.9675\\
				&\multicolumn{1}{l|}{COAST \cite{you2021coast}} &22.45/0.6800&25.94/0.8035&31.10/0.9168&32.23/0.9321&34.22/0.9530&35.99/0.9665\\  \cline{2-8}
				&\multicolumn{1}{l|}{\multirow{2}{*}{DP-DUN}} &\textcolor{red}{23.61/0.7268}&\textcolor{red}{27.03}/\textcolor{blue}{0.8335}&\textcolor{red}{32.43}/\textcolor{red}{0.9328}&\textcolor{red}{33.70}/\textcolor{red}{0.9464}&\textcolor{red}{35.69}/\textcolor{red}{0.9628}&\textcolor{red}{37.63}/\textcolor{red}{0.9743}\\
				&\multicolumn{1}{c|}{}&\textcolor{cyan}{(23.0/25.0)}&\textcolor{cyan}{(22.9/23.3)}&\textcolor{cyan}{(19.0/21.1)}&\textcolor{cyan}{(21.0/22.8)}&\textcolor{cyan}{(18.9/21.9)}&\textcolor{cyan}{(17.0/21.2)}\\
				&\multicolumn{1}{l|}{\multirow{2}{*}{DPC-DUN}} &\textcolor{blue}{23.43}/\textcolor{blue}{0.7177}&\textcolor{blue}{26.99}/\textcolor{red}{0.8345}&\textcolor{blue}{32.36}/\textcolor{blue}{0.9323}&\textcolor{blue}{33.53}/\textcolor{blue}{0.9449}&\textcolor{blue}{35.61}/\textcolor{blue}{0.9624}&\textcolor{blue}{37.52}/\textcolor{blue}{0.9737}\\
				&\multicolumn{1}{c|}{}&\textcolor{cyan}{(16.0/17.4)}&\textcolor{cyan}{(20.5/21.7)}&\textcolor{cyan}{(18.2/19.5)}&\textcolor{cyan}{(17.3/18.5)}&\textcolor{cyan}{(16.3/18.3)}&\textcolor{cyan}{(14.7/17.5)}\\
				\toprule
				\multirow{7}{*}{DIV2K}
				&\multicolumn{1}{l|}{ISTA-Net$^+$ \cite{zhang2018ista}} &25.15/0.6922&27.73/0.7859&32.24/0.9007&33.43/0.9201&35.43/0.9450&37.33/0.9617\\
				&\multicolumn{1}{l|}{iPiano-Net \cite{su2020ipiano}} &26.38/0.7410&28.94/0.8232&33.29/0.9164&34.47/0.9326&36.50/0.9539&\textcolor{blue}{38.36}/0.9677\\
				&\multicolumn{1}{l|}{COAST \cite{you2021coast}} &26.37/0.7464&29.17/0.8276&33.45/\textcolor{blue}{0.9178}&34.49/0.9323&36.41/0.9530&38.22/0.9668\\  \cline{2-8}
				&\multicolumn{1}{l|}{\multirow{2}{*}{DP-DUN}} &\textcolor{red}{27.07}/\textcolor{red}{0.7630}&\textcolor{red}{29.70}/\textcolor{blue}{0.8401}&\textcolor{red}{34.13}/\textcolor{red}{0.9257}&\textcolor{red}{35.29}/\textcolor{red}{0.9403}&\textcolor{red}{37.27}/\textcolor{red}{0.9591}&\textcolor{red}{39.11}/\textcolor{blue}{0.9711}\\
				&\multicolumn{1}{c|}{}&\textcolor{cyan}{(23.0/25.0)}&\textcolor{cyan}{(23.0/23.3)}&\textcolor{cyan}{(19.0/20.7)}&\textcolor{cyan}{(21.0/22.4)}&\textcolor{cyan}{(18.7/21.1)}&\textcolor{cyan}{(17.0/20.2)}\\
				&\multicolumn{1}{l|}{\multirow{2}{*}{DPC-DUN}} &\textcolor{blue}{26.98}/\textcolor{blue}{0.7626}&\textcolor{blue}{29.67}/\textcolor{red}{0.8423}&\textcolor{blue}{34.08}/\textcolor{red}{0.9257}&\textcolor{blue}{35.20}/\textcolor{blue}{0.9397}&\textcolor{blue}{37.20}/\textcolor{blue}{0.9588}&\textcolor{red}{39.11}/\textcolor{red}{0.9713}\\
				&\multicolumn{1}{c|}{}&\textcolor{cyan}{(15.8/17.5)}&\textcolor{cyan}{(19.4/21.2)}&\textcolor{cyan}{(17.5/18.8)}&\textcolor{cyan}{(16.4/18.0)}&\textcolor{cyan}{(15.4/17.5)}&\textcolor{cyan}{(14.1/16.7)}\\
				\toprule
		\end{tabular}}
		\vspace{-0.1cm}
	\end{table*}
%%%%%%%%%%%%%%%%%%%%%%%%%%%%%%%%%%%%
%%%%%%%%%%%%%%%%%%%%%%%%%%%%%%%%%%%%%%%%
\begin{figure*}[t]
\centering
\includegraphics[width=1.0\textwidth]{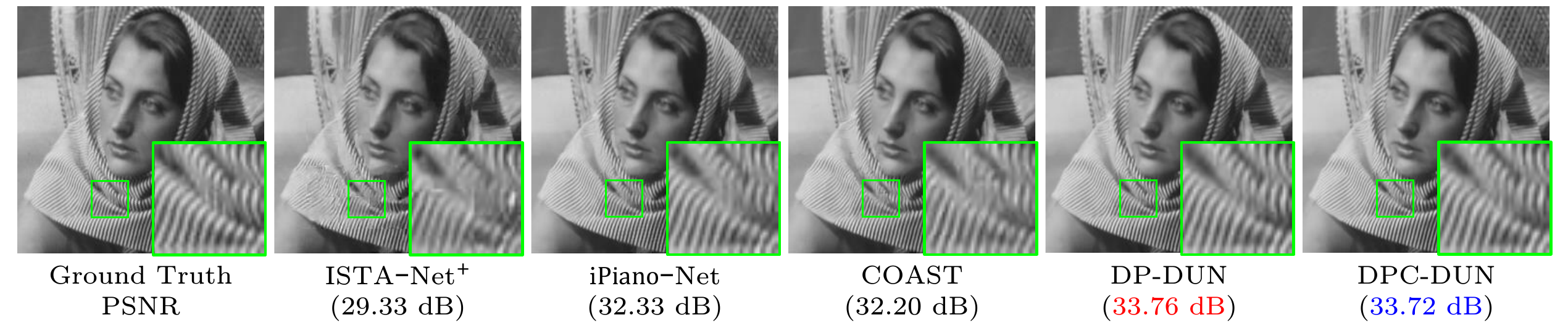} 
\vspace{-0.7cm}
\caption{Comparisons on recovering an image (``Barbara'') from Set11 dataset \cite{Kulkarni2016ReconNetNR} in the case of CS ratio = 25$\%$.}
\vspace{-0.2cm}
\label{fig:Set11_CBSD68}
\end{figure*}
%%%%%%%%%%%%%%%%%%%%%%%%%%%%%%%%%%%%%%%%
%%%%%%%%%%%%%%%%%%%%%%%%%%%%%%%%%%%%%%%%
\begin{figure*}[t]
\centering
\setlength{\abovecaptionskip}{-1pt}
\setlength{\belowcaptionskip}{-0.cm}
\includegraphics[width=1.0\textwidth]{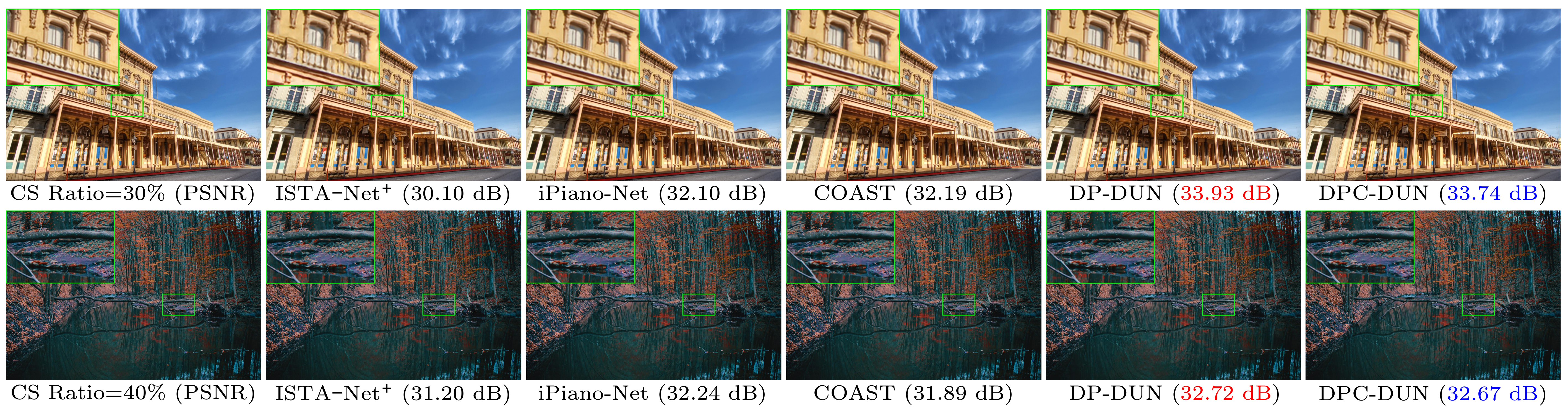} 
% \vspace{-0.6cm}
\caption{Comparisons on an image from Urban100 dataset \cite{dong2018denoising} in the case of CS ratio = 30\% (upper) and an image from DIV2K dataset \cite{Agustsson_2017_CVPR_Workshops} in the case of CS ratio = 40\% (lower).}
\vspace{-0.3cm}
\label{fig:Urban100_DIV2K}
\end{figure*}
%%%%%%%%%%%%%%%%%%%%%%%%%%%%%%%%%%%%%%%%
%%%%%%%%%%%%%%%%%%%%%%%%%%%%%%%%%%%%%%%%
\begin{figure*}[t]
\centering
\setlength{\abovecaptionskip}{0.cm}
\setlength{\belowcaptionskip}{-0.cm}
\includegraphics[width=1.0\textwidth]{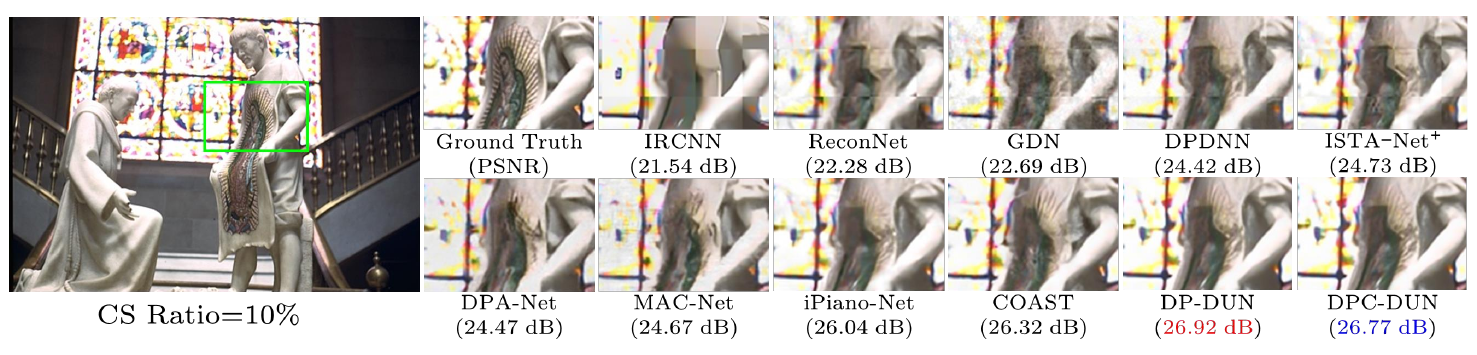} 
\vspace{-0.6cm}
\caption{Comparisons on recovering an image from CBSD68 dataset \cite{zhang2021plug} in the case of CS ratio = 10\%.}
\vspace{-0.2cm}
\label{fig:Set11_25}
\end{figure*}
%%%%%%%%%%%%%%%%%%%%%%%%%%%%%%%%%%%%%%%%
%%%%%%%%%%%%%%%%%%%%%%%%%%%%%%%%%%%%
\begin{table}[t]
		\centering
		\setlength{\abovecaptionskip}{0pt}
        \setlength{\belowcaptionskip}{0pt}
        \large
% 		\small
% 		\caption{The computational cost (FLOPs) and the parameter number (Paras) comparisons on different modules.}
		\caption{The computation cost and parameter number comparisons of all modules.}
		\label{tab:para}
% 		\vspace{5pt}
        % \resizebox{\textwidth}{!}{%
        % \renewcommand\arraystretch{0.01}
		\resizebox{8.5cm}{!}{%
		\renewcommand
% 		\arraystretch{0.01}
		\tabcolsep{10pt}
			\begin{tabular}{c|cccc}
			\toprule
				% \hline
				\multicolumn{1}{c|}{} & \multicolumn{1}{c}{GDM} & \multicolumn{1}{c}{PMM} & \multicolumn{1}{c}{CU} & \multicolumn{1}{c}{PS} \\
				% \cline{1-5}
				\hline
				\multicolumn{1}{l|}{FLOPs} & $9.1 \times 10^{7}$ & $7.7 \times 10^{9}$ & $1.3 \times 10^{9}$ & $5.6 \times 10^{2}$ \\
				% \cline{1-5}
				\multicolumn{1}{l|}{Paras} & 1 & 55424 & 11296 & 292 \\
				\toprule
		\end{tabular}}
		\vspace{-0.7cm}
	\end{table}
%%%%%%%%%%%%%%%%%%%%%%%%%%%%%%%%%%%%
%%%%%%%%%%%%%%%%%%%%%%%%%%%%%%%%%%%%%%%%
\begin{figure*}[!t]
\centering
\vspace{-0.1cm}
\includegraphics[width=1.0\textwidth]{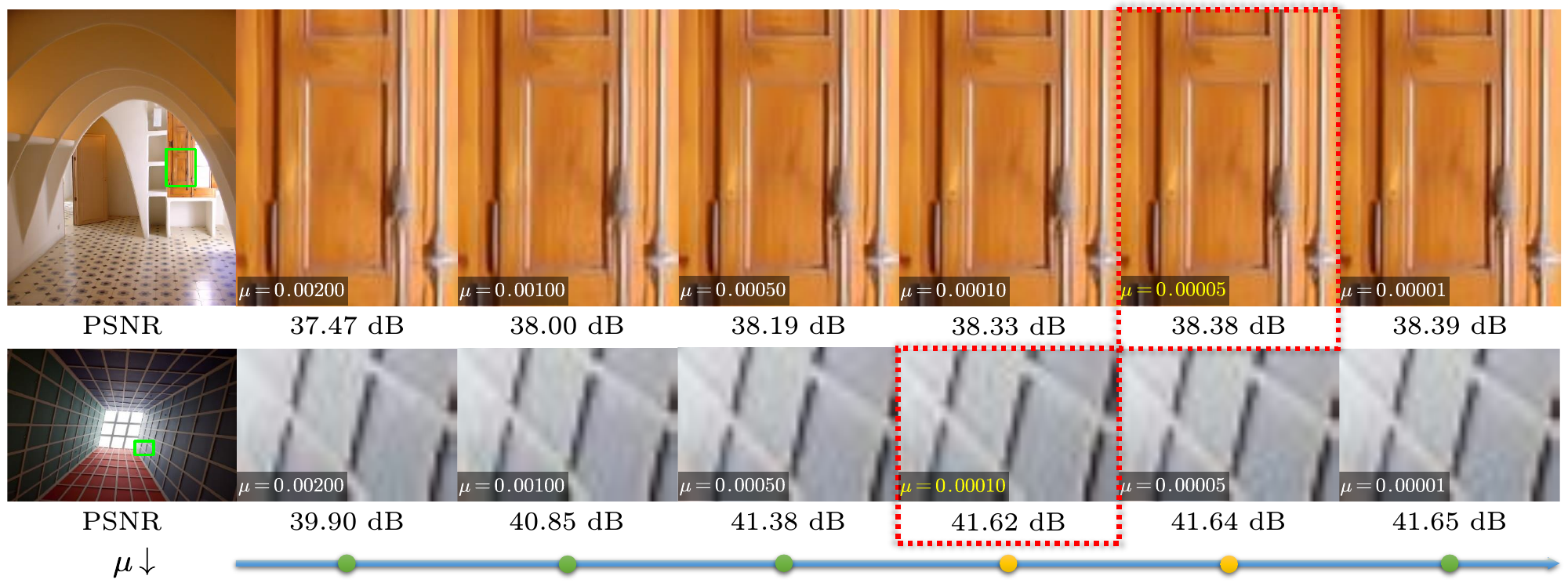}
\vspace{-0.5cm}
\caption{Illustration of recovering two images from Urban100 dataset \cite{dong2018denoising} when CS ratio = $10\%$ under different $\mu$ with one single DPC-DUN model. The recovery accuracies and computational costs stably increase and converge with a gradual reduction of $\mu$. Users in various scenarios can adjust the suppression intensity on resource burden by a sliding bar (in \textcolor{blue}{blue}) to find satisfactory tradeoffs with higher accuracy and lower cost, and the ideal settings are marked in \textcolor{red}{red} boxes with \textcolor{yellow}{yellow} setting color.}
\vspace{-0.2cm}
\label{fig:adjust_comp}
\end{figure*}
%%%%%%%%%%%%%%%%%%%%%%%%%%%%%%%%%%%%%%%%
%%%%%%%%%%%%%%%%%%%%%%%%%%%%%%%%%%%%
\begin{table*}[t]
		\centering
		\setlength{\abovecaptionskip}{0pt}
        \setlength{\belowcaptionskip}{0pt}
		\small
		\caption{Average PSNR(dB)/FLOPs($10^9$)/$N_{AM}$ performance comparisons on Set11 \cite{Kulkarni2016ReconNetNR} and CBSD68 \cite{zhang2021plug} datasets with different stage number when CS ratio is $30\%$. After using the PS or PCS, the PSNR value remains unchanged, and the computation FLOPs drops more obviously with the larger number of stages.  
		}
		\label{tab:stage_num}
% 		\vspace{5pt}
		\resizebox{\textwidth}{!}{%
		\renewcommand\tabcolsep{8pt}
			\begin{tabular}{cc|cccc|cccc}
			\toprule
				% \hline
				\multicolumn{2}{c|}{\multirow{2}{*}{Stage Number}} & \multicolumn{4}{c|}{Set11} & \multicolumn{4}{c}{CBSD68} \\ \cline{3-10}
				\multicolumn{1}{c}{}&\multicolumn{1}{c|}{}&\multicolumn{1}{c}{20}&\multicolumn{1}{c}{25}&\multicolumn{1}{c}{30}&\multicolumn{1}{c|}{35}&\multicolumn{1}{c}{20}&\multicolumn{1}{c}{25}&\multicolumn{1}{c}{30}&\multicolumn{1}{c}{35}\\ 
				% \cline{3-7}
				% &&10\%&25\%&30\%&40\%&50\%\\
				\toprule
				% \hline
				% \hline
				\multicolumn{1}{c|}{\multirow{2}{*}{w/o PS}}&\multicolumn{1}{l|}{PSNR(dB)} &35.98&36.05&36.07&36.10&31.81&31.85&31.86&31.87\\
				\multicolumn{1}{c|}{}&\multicolumn{1}{l|}{FLOPs($10^9$)} &155.9&193.6&232.2&271.0&363.0&453.7&544.4&635.1\\ \cline{1-10}
				
				\multicolumn{1}{c|}{\multirow{3}{*}{w/ PS}}&\multicolumn{1}{l|}{PSNR(dB)} &35.99&36.02&36.06&36.08&31.82&31.83&31.84&31.86\\
				\multicolumn{1}{c|}{}&\multicolumn{1}{l|}{FLOPs($10^9$)} &154.9&178.2&186.1&238.9&361.0&414.0&433.2&552.9\\ 
				\multicolumn{1}{c|}{}&\multicolumn{1}{l|}{$N_{AM}$} &18.0/19.9&21.0/22.9&22.0/23.9&26.1/30.7&18.0/19.8&21.0/22.7&22.0/23.8&26.0/30.3\\
				\cline{1-10}
				
				\multicolumn{1}{c|}{\multirow{3}{*}{w/ PCS}}&\multicolumn{1}{l|}{PSNR(dB)} &35.89&35.88&35.93&35.98&31.76&31.76&31.79&31.79\\
				\multicolumn{1}{c|}{}&\multicolumn{1}{l|}{FLOPs($10^9$)} &167.8&174.5&182.2&206.2&389.4&412.8&460.5&490.5\\ 
				\multicolumn{1}{c|}{}&\multicolumn{1}{l|}{$N_{AM}$} &16.0/18.3&16.5/18.3&17.5/19.5&18.1/20.7&15.6/18.1&17.2/18.5&17.7/20.3&17.9/21.1\\
				\toprule
				% \hline
				% \hline
		\end{tabular}}
		\vspace{-0.5cm}
	\end{table*}
%%%%%%%%%%%%%%%%%%%%%%%%%%%%%%%%%%%%
% %%%%%%%%%%88%%%%%%%%%%%%%%%%%%
\begin{table}[!t]
% \vspace{-12pt}
\centering
\small
% \normalsize
% \footnotesize 
\setlength{\abovecaptionskip}{0pt}
\setlength{\belowcaptionskip}{0pt}
\caption{The Memory/FLOPs/PSNR performance on Set11 dataset.} 
\label{tab:comparision}
\renewcommand\tabcolsep{5pt}
\begin{tabular}{ccccc}
\toprule
% \hline
\multicolumn{1}{c|}{Method} & \multicolumn{1}{c}{iPiano-Net} & \multicolumn{1}{c}{COAST} & \multicolumn{1}{c}{DP-DUN} & \multicolumn{1}{c}{DPC-DUN}
% \\\multicolumn{1}{c|}{} &  \multicolumn{1}{c}{} &  \multicolumn{1}{c}{} & \multicolumn{1}{c}{($\mu=0.002$)} & \multicolumn{1}{c}{($\mu=0.002$)}
\\ \hline  
\multicolumn{1}{c|}{Mem.(MB)}  & \multicolumn{1}{c}{1810} & \multicolumn{1}{c}{1544} & \multicolumn{1}{c}{1312} & \multicolumn{1}{c}{1442}
\\
\multicolumn{1}{c|}{FLOPs($10^9$)}  & \multicolumn{1}{c}{1426.7} & \multicolumn{1}{c}{157.7} & \multicolumn{1}{c}{72.0} & \multicolumn{1}{c}{138.0}
\\
\multicolumn{1}{c|}{PSNR(dB)}  & \multicolumn{1}{c}{34.78} & \multicolumn{1}{c}{35.11} & \multicolumn{1}{c}{35.34} & \multicolumn{1}{c}{35.54}
\\
\toprule
% \hline
\end{tabular}
\vspace{-0.5cm}
% \vspace{-12pt}
\end{table}
%%%%%%%%%%%%%%%%%%%%%%%%%%%%%%%%%%%%%%%%

We use the 400 training images of size $180 \times 180$ \cite{chen2016trainable}, generating the training data pairs $\{(\mathbf{y}_j, \mathbf{x}_j)\}_{j=1}^{N_a}$ by extracting the luminance component of each image block of size $33 \times 33$, \textit{i.e.} $N=1,089$. Meanwhile, we apply the data augmentation technique to increase the data diversity. For a given CS ratio, the corresponding measurement matrix $\mathbf{\Phi}\in\mathbb{R}^{M \times N}$ is constructed by generating a random Gaussian matrix and
then orthogonalizing its rows, \textit{i.e.} $\mathbf{\Phi}\mathbf{\Phi^{\top}} = \mathbf{I}$, where $\mathbf{I}$ is the identity matrix. The sampling process applies a convolution layer whose kernel is the matrix $\mathbf{\Phi}$. Applying $\mathbf{y}_j=\mathbf{\Phi}\mathbf{x}_j$ yields the set of CS measurements, where $\mathbf{x}_j$ is the vectorized version of an image block and is initialized by $\mathbf{x}_j^{(0)}=\mathbf{\Phi^{\top}}\mathbf{y}_j$.

Our proposed models are trained with 410 epochs separately for each CS ratio. Each image block of size $33 \times 33$ is sampled and reconstructed independently for the first 400 epochs, and for the last ten epochs, we adopt larger image blocks of size $99 \times 99$ as the inputs to finetune the model further. To alleviate blocking artifacts, we firstly unfold the blocks of size $99 \times 99$ into overlapping blocks of size $33 \times 33$ while sampling process $\mathbf{\Phi}\mathbf{x}$ and then fold the blocks of size $33 \times 33$ into larger blocks while initialization $\mathbf{\Phi^\top}\mathbf{y}$ \cite{su2020ipiano}. We also unfold the whole image with this approach during testing. We use momentum of 0.9 and weight decay of 0.999. The default batch size is 64, the default stage number $K$ is 25, the default number of feature maps $C$ is 32, and the learnable parameter $\rho^{(k)}$ is initialized to 1. 
% The default Lagrange multiplier $\mu$ is 0.00005 in DP-DUN, and 
In the training phase, the default Lagrange multiplier $\mu$ is randomly chosen as the input of DPC-DUN from the range of $\{0.00001, 0.00005, 0.0001, 0.0005, 0.001, 0.002\}$. The CS reconstruction accuracies on all datasets are evaluated with PSNR and SSIM. And we also show the active module number ($N_{AM}$) and the computation cost (including the computations of the convolution and the fully-connected layer) measured in floating-point operations per second (FLOPs).

\subsection{Qualitative Evaluation}

We compare our proposed DP-DUN ($\mu=0.00005$) and DPC-DUN ($\mu=0.00001$) with some recent representative CS reconstruction methods. The average PSNR/SSIM reconstruction performances on Set11 \cite{Kulkarni2016ReconNetNR} and CBSD68 \cite{zhang2021plug} datasets with respect to five CS ratios are summarized in Table~\ref{tab:set11} and Table~\ref{tab:cbsd68} respectively. Furthermore, Table~\ref{tab:urban100_div2k} shows the comparisons with respect to six CS ratios on Urban100 \cite{dong2018denoising} and DIV2K \cite{Agustsson_2017_CVPR_Workshops} datasets which contain more and larger images. One can observe that our DP-DUN and DPC-DUN outperform all the other competing methods in PSNR across all the cases. Compared with SCS-GNet \cite{zhong2022scalable}, our methods achieve a high PSNR and slightly lower SSIM for some ratios in Table~\ref{tab:set11}. However, SCS-GNet takes more than half an hour to reconstruct an image of size 256 $\times$ 256 for testing, while our methods achieve real-time performance. So our methods can save resources while maintaining high performance. Therefore, our model applies well to the fixed Gaussian matrix and can be easily extended to other deterministic measurement matrices. 
% \cite{lu2017binary}. 
In addition, the number of active modules decreases significantly, especially when the CS ratio is larger, and DPC-DUN can utilize the smaller numbers of GDM and PMM than DP-DUN in a less volatile range of the PSNR value. Fig.~\ref{fig:Set11_CBSD68}, Fig.~\ref{fig:Urban100_DIV2K}, and Fig.~\ref{fig:Set11_25} further show the visual comparisons of challenging images on various datasets. Our DP-DUN and DPC-DUN generate visually pleasant images that are faithful to the ground truth. 

\subsection{Cost Comparison}

%%%%%%%%%%%%%%%%%%%%%%%%%%%%%%%%%%%%%%%%
\begin{figure*}[!t]
\centering
\setlength{\abovecaptionskip}{0.cm}
\setlength{\belowcaptionskip}{-0.cm}
\includegraphics[width=1.0\textwidth]{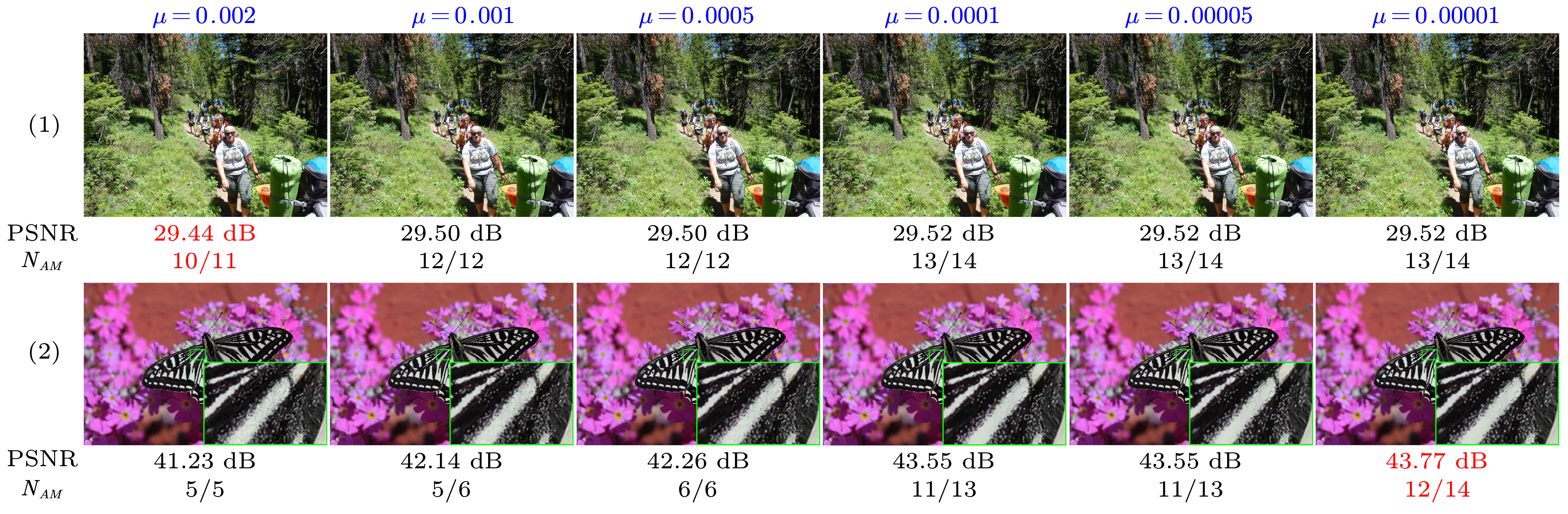} 
\vspace{-0.4cm}
\caption{Illustration of recovering two images from DIV2K dataset \cite{Agustsson_2017_CVPR_Workshops} when CS ratio = $50\%$ under different $\mu$ with one single DPC-DUN model. The PSNR value and the active module numbers ($N_{AM}$) of GDM/PMM highlighted in \textcolor{red}{red} represent the optimal choice.}
\vspace{-0.2cm}
\label{fig:dpcdun}
\end{figure*}
%%%%%%%%%%%%%%%%%%%%%%%%%%%%%%%%%%%%%%%%
%%%%%%%%%%%%%%%%%%%%%%%%%%%%%%%%%%%%%%%%
\begin{figure*}[!t]
\centering
\setlength{\abovecaptionskip}{0pt}
\setlength{\belowcaptionskip}{0cm}
\includegraphics[width=0.95\textwidth]{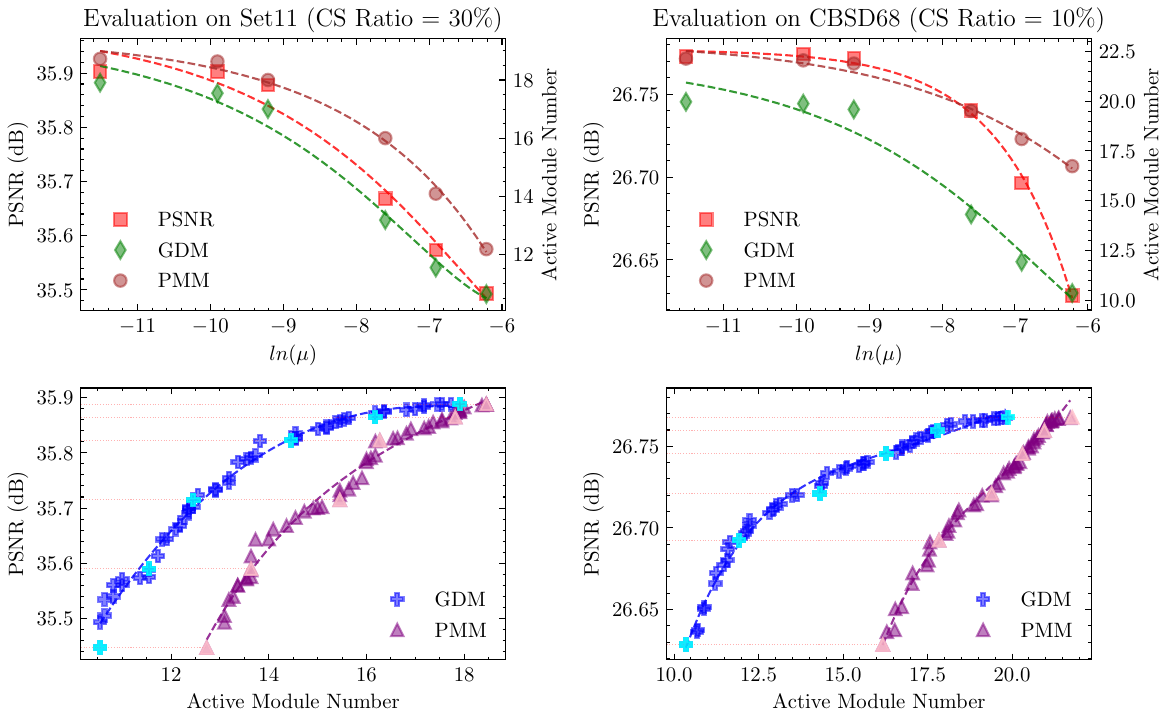}
\vspace{-4pt}
\caption{Visualization of the relationships among the factor $\mu$, the active module numbers of GDM/PMM, and PSNR on Set11 dataset \cite{Kulkarni2016ReconNetNR} with CS ratio = 30$\%$ (left) and CBSD68 dataset \cite{zhang2021plug} with CS ratio = 10\% (right). The upper row shows the stable accuracy and computational cost reductions with the increase of $\mu$, and their inconsistent changing degrees under different settings. The lower row exhibits the scalability and robustness of DPC-DUN, which can flexibly generalize to unseen configurations by learning with only six default settings (highlighted in light colors with \textcolor{orange}{orange} dotted lines).}
\vspace{-0.5cm}
\label{fig:eval_mu}
\end{figure*}
%%%%%%%%%%%%%%%%%%%%%%%%%%%%%%%%%%%%%%%%

We analyze the FLOPs on Set11 dataset and the parameter numbers of each module in Table~\ref{tab:para}. Compared with others, the lightweight path selector (PS) requires less storage space and computational overhead, which can help reduce the memory and cost of the network. What is more, Table~\ref{tab:stage_num} compares w/o PS,  w/ PS and w/ PCS with different stage numbers when CS ratio is $30\%$. It is clear that the PSNR value remains unchanged when stage number $K \geqslant 25$, and although adding PS increases the computation costs of PS, the numbers of total FLOPs are reduced, especially when the stage number is much larger. Path-controllable selector (PCS), which contains PS and CU, gives the network less computational overhead than only using PS while the performance changes little, especially $K \geqslant 25$. Therefore, CU containing a convolution layer has more parameters but can reduce total model numbers compared with the non-adjustable model.

The GPU memory and FLOPs of some competing methods are provided in Table~\ref{tab:comparision}. Obviously, our DP-DUN/DPC-DUN ($\mu=0.002$) has less memory and computation cost than others while ensuring the best performance. This demonstrates that our methods can achieve the best performance-consumption balance by adjusting the multiplier.

\subsection{Analysis of Path-Controllable Selector (PCS)}

%%%%%%%%%%%%%%%%%%%%%%%%%%%%%%%%%%%%
\begin{table*}[!t]
		\centering
		\setlength{\abovecaptionskip}{0pt}
        \setlength{\belowcaptionskip}{2pt}
		\small
		\caption{Average performance comparisons on Set11 \cite{Kulkarni2016ReconNetNR} and CBSD68 \cite{zhang2021plug} datasets with various Lagrange multipliers when CS ratio is 30$\%$. The larger the value of $\mu$, the smaller the value of PSNR, the smaller the value of $N_{AM}$, and the lower the computation cost (FLOPs).}
		\label{tab:mu}
% 		\vspace{5pt}
		\resizebox{\textwidth}{!}{%
		\renewcommand\tabcolsep{8pt}
			\begin{tabular}{cc|cccccccc}
			\toprule
				% \hline
				\multicolumn{2}{c|}{\multirow{1}{*}{$\mu$}} &0&0.00001&0.00005&0.0001&0.0002&0.0005&0.001&0.002\\ 
				% \cline{3-7}
				% &&10\%&25\%&30\%&40\%&50\%\\
				\toprule
				% \hline
				% \hline
				\multicolumn{1}{c|}{\multirow{3}{*}{Set11}}&\multicolumn{1}{l|}{PSNR(dB)} &36.05&36.04&36.02&36.04&35.98&35.84&35.69&35.34\\
				\multicolumn{1}{c|}{}&\multicolumn{1}{l|}{FLOPs($10^9$)} &194.0&185.9&178.2&182.0&157.9&116.6&98.7&72.4\\ 
				\multicolumn{1}{c|}{}&\multicolumn{1}{l|}{$N_{AM}$} &25.0/24.9&21.0/23.9&21.0/22.9&21.0/23.4&17.5/20.3&12.4/15.0&9.6/12.7&8.2/9.3\\ 
				\cline{1-10}
				
				\multicolumn{1}{c|}{\multirow{3}{*}{CBSD68}}&\multicolumn{1}{l|}{PSNR(dB)} &31.84&31.83&31.83&31.83&31.81&31.73&31.62&31.38\\
				\multicolumn{1}{c|}{}&\multicolumn{1}{l|}{FLOPs($10^9$)} &452.8&430.3&414.0&426.6&368.1&284.1&220.3&164.1\\ 
				\multicolumn{1}{c|}{}&\multicolumn{1}{l|}{$N_{AM}$} &25.0/24.8&21.0/23.6&21.0/22.7&20.9/23.4&17.1/20.2&12.0/15.6&9.0/12.1&7.6/9.0\\
				\toprule
				% \hline
				% \hline
		\end{tabular}}
		\vspace{-0.5cm}
	\end{table*}
%%%%%%%%%%%%%%%%%%%%%%%%%%%%%%%%%%%%

By combining the information of a one-hot vector encoded by the Lagrange multiplier $\mu$ and the input image feature, DPC-DUN can dynamically balance the recovery accuracy and resource overhead with our proposed controllable mechanism achieved by PCS. In practice, users can interactively adjust the tradeoff of performance-complexity by manipulating a sliding bar, and find property values for specific scenarios. Fig.~\ref{fig:adjust_comp} presents our recoveries under different $\mu$ values and exhibits the customized reconstructions of DPC-DUN with stable performance. 
% The value of PSNR increases with the reduction of penalty factor $\mu$. 
To more fully show the role of the modulation mechanism for the specific images, we present more qualitative results from DIV2K dataset \cite{Agustsson_2017_CVPR_Workshops} under different $\mu$ in Fig.~\ref{fig:dpcdun}. We find that with the increase of the $\mu$ value, the PSNR value increases insignificantly, and yet the number of the active module ($N_{AM}$) increases obviously in Fig.~\ref{fig:dpcdun}(1), so $\mu=0.002$ is the best choice to achieve the performance-complexity tradeoff for the easier image. For the image of Fig.~\ref{fig:dpcdun}(2) which contains rich contents, the performance and the computational complexity are influenced significantly by the value of $\mu$, and we select the setting $\mu=0.00001$ to maintain great reconstruction accuracy. Therefore, our DPC-DUN is designed more finely and specifically by controlling the value of $\mu$.

To analyze the relationships among $\mu$, active module numbers of GDM/PMM and PSNR, Fig.~\ref{fig:eval_mu} shows the changes through evaluations on Set11 \cite{Kulkarni2016ReconNetNR} and CBSD68 \cite{zhang2021plug} datasets. 
% From the upper subplot, we can see that with the increase of penalty factor $\mu$, the value of PSNR and the active module numbers both decrease. 
From the upper subplot, we can see that PSNR and active module numbers generally keep consistent trends but maybe with different changing degrees under various sampling rates and data distributions. Moreover, to break the limits of the fixed one-hot vectors encoded by the modulation factor $\mu$, we explore the effect of other (\emph{unseen}) binary encodings in our proposed method, as shown in the lower row of Fig.~\ref{fig:eval_mu}. In addition to the provided six default settings adopted in training, DPC-DUN can generalize to unseen encodings, enabling fine-granular model scalability. These facts verify our controllable and adaptive design's good interactivity and high flexibility.

%%%%%%%%%%%%%%%%%%%%%%%%%%%%%%%%%%%%
\begin{table}[t]
		\centering
		\setlength{\abovecaptionskip}{0pt}
        \setlength{\belowcaptionskip}{0pt}
		\small
		\caption{Average PSNR(dB)/SSIM/$N_{AM}$ performance results of DP-DUN/DPC-DUN on Set11 dataset, where ``-'' represents the results without fine-tuning (deblocking).}
		\label{tab:deblock}
% 		\vspace{5pt}
        \scalebox{1}{
% 		\resizebox{\textwidth}{!}{%
		\renewcommand\tabcolsep{3pt}
			\begin{tabular}{c|cc|cc}
			\toprule
				% \hline
				\multicolumn{1}{c|}{\multirow{2}{*}{CS Ratio}} & \multicolumn{2}{c|}{10} & \multicolumn{2}{c}{30} \\ \cline{2-5}
				\multicolumn{1}{c|}{}&\multicolumn{1}{c}{PSNR/SSIM}&\multicolumn{1}{c|}{$N_{AM}$}&\multicolumn{1}{c}{PSNR/SSIM}&\multicolumn{1}{c}{$N_{AM}$}\\ 
				% \cline{3-7}
				% &&10\%&25\%&30\%&40\%&50\%\\
				\toprule
				% \hline
				% \hline
				\multicolumn{1}{l|}{DP-DUN$^{-}$}&29.20/0.8737&23.0/23.0&35.93/0.9569&21.0/22.8\\
                \multicolumn{1}{l|}{DP-DUN}&29.42/0.8806&22.9/23.2&36.02/0.9577&21.0/22.9\\
                \hline
                \multicolumn{1}{l|}{DPC-DUN$^{-}$}&29.17/0.8740&20.3/21.4&35.69/0.9560&15.7/16.9\\
                \multicolumn{1}{l|}{DPC-DUN}&29.40/0.8798&20.5/21.6&35.88/0.9570&16.5/18.3\\
				\toprule
				% \hline
				% \hline
		\end{tabular}}
% 		}
		\vspace{-0.4cm}
	\end{table}
%%%%%%%%%%%%%%%%%%%%%%%%%%%%%%%%%%%%
%%%%%%%%%%%%%%%%%%%%%%%%%%%%%%%%%%%%
\begin{table}[t]
		\centering
		\setlength{\abovecaptionskip}{0pt}
        \setlength{\belowcaptionskip}{0pt}
		\small
		\caption{Average PSNR(dB)/SSIM performance comparisons on Set11 dataset with learnable sampling matrix. The best performance is labeled in \textbf{bold}.}
		\label{tab:learned_comp}
% 		\vspace{5pt}
        \scalebox{1}{
% 		\resizebox{\textwidth}{!}{%
		\renewcommand\tabcolsep{7pt}
			\begin{tabular}{c|cc|cc}
			\toprule
				% \hline
				\multicolumn{1}{c|}{\multirow{2}{*}{CS Ratio}} & \multicolumn{2}{c|}{10} & \multicolumn{2}{c}{30} \\ \cline{2-5}
				\multicolumn{1}{c|}{}&\multicolumn{1}{c}{PSNR}&\multicolumn{1}{c|}{SSIM}&\multicolumn{1}{c}{PSNR}&\multicolumn{1}{c}{SSIM}\\ 
				% \cline{3-7}
				% &&10\%&25\%&30\%&40\%&50\%\\
				\toprule
				% \hline
                \multicolumn{1}{l|}{\multirow{1}{*}{MADUN \cite{song2021memory}}} 
                &29.91&0.8986&36.94&0.9676\\
                \multicolumn{1}{l|}{\multirow{1}{*}{HQSRED-Net \cite{ma2022deep}}} 
                &29.04&0.8678&35.58&0.9553\\
                \multicolumn{1}{l|}{\multirow{1}{*}{SCS-GNet \cite{zhong2022scalable}}} 
                &29.35&0.8854&35.42&0.9588\\
                \multicolumn{1}{l|}{\multirow{1}{*}{TransCS \cite{shen2022transcs}}} 
                &29.54&0.8877&35.62&0.9588\\ \hline
				\multicolumn{1}{l|}{\multirow{2}{*}{DPC-DUN}} 
                &\textbf{30.00}&\textbf{0.9002}&\textbf{37.02}&\textbf{0.9677}\\
                \multicolumn{1}{c|}{}&\multicolumn{2}{c|}{\textcolor{cyan}{(13.0/24.0)}}&\multicolumn{2}{c}{\textcolor{cyan}{(12.9/23.5)}}\\
				\toprule
				% \hline
				% \hline
		\end{tabular}}
% 		}
		\vspace{-0.5cm}
	\end{table}
%%%%%%%%%%%%%%%%%%%%%%%%%%%%%%%%%%%%

\subsection{Analysis of Path Selector (PS)}\label{sec:mu}
% %%%%%%%%%%%%%%%%%%%%%%%%%%%%%%%%%%%%
% \begin{table}[t]
% 		\centering
% 		\setlength{\abovecaptionskip}{0pt}
%         \setlength{\belowcaptionskip}{5pt}
% % 		\small
%         % \renewcommand{\thetable}{\textbf{VII}}
% 		\caption{Average PSNR(dB)/SSIM/$N_{AM}$ performance results of DP-DUN/DPC-DUN on Set11 dataset with the different sampling matrix.}
% 		\label{tab:other_matrix}
% % 		\vspace{5pt}
%         \scalebox{1}{
% % 		\resizebox{\textwidth}{!}{%
% 		\renewcommand\tabcolsep{3pt}
% 			\begin{tabular}{cc|cc|cc}
% 			\toprule
% 				% \hline
% 				\multicolumn{2}{c|}{\multirow{2}{*}{CS Ratio}} & \multicolumn{2}{c|}{10} & \multicolumn{2}{c}{30} \\ \cline{3-6}
% 				\multicolumn{2}{c|}{}&\multicolumn{1}{c}{PSNR/SSIM}&\multicolumn{1}{c|}{$N_{AM}$}&\multicolumn{1}{c}{PSNR/SSIM}&\multicolumn{1}{c}{$N_{AM}$}\\ 
% 				% \cline{3-7}
% 				% &&10\%&25\%&30\%&40\%&50\%\\
% 				\toprule
% 				% \hline
% 				% \hline
% 				\multicolumn{1}{c|}{\multirow{2}{*}{Learned}} &\multicolumn{1}{c|}{DP-DUN}&30.16/0.9015&23.0/25.0&36.79/0.9674&23.2/25.0\\
%                 \multicolumn{1}{c|}{}&\multicolumn{1}{c|}{DPC-DUN}&30.00/0.9002&13.0/24.0&37.02/0.9677&12.9/23.5\\
%                 \hline
%                 \multicolumn{1}{c|}{\multirow{2}{*}{Binary}}&\multicolumn{1}{c|}{DP-DUN}&30.36/0.9011&17.8/17.0&37.20/0.9667&20.9/19.6\\
%                 \multicolumn{1}{c|}{}&\multicolumn{1}{c|}{DPC-DUN}&30.30/0.9003&12.5/12.1&37.02/0.9658&15.0/11.0\\
% 				\toprule
% 				% \hline
% 				% \hline
% 		\end{tabular}}
% % 		}
% 		\vspace{-0.2cm}
% 	\end{table}
% %%%%%%%%%%%%%%%%%%%%%%%%%%%%%%%%%%%%

In our proposed methods, path selector (PS) plays a significant role in minimizing the computation burden at test time, which is important for optimizing models to meet practical application needs. Lagrange multiplier $\mu$ in Eq.~\eqref{eq: loss_total} represents the factor that selects a specific performance-complexity tradeoff point. We conduct experiments on Set11 \cite{Kulkarni2016ReconNetNR} and CBSD68 \cite{zhang2021plug} datasets to analyze the impact of different values of $\mu$ on different tradeoff, as shown in Table~\ref{tab:mu}. The greater the value of $\mu$, the greater the weight of the model computation cost in the optimization problem, thus, the PSNR value is smaller, $N_{AM}$ and FLOPs are also smaller. Also, the PSNR value does not change obviously when $\mu \leqslant 0.0002$ and compared with $\mu=0$, the FLOPs on $\mu=0.0002$ is reduced about $18.2\%$ on Set11 dataset and about $18.6\%$ on CBSD68 dataset, which shows that adjusting PS can save computing resources extremely well.

\subsection{Ablation Study}

%%%%%%%%%%%%%%%%%%%%%%%%%%%%%%%%%%%
\begin{table*}[t]
\centering
\setlength{\abovecaptionskip}{0pt}
\setlength{\belowcaptionskip}{5pt}
\small
% \renewcommand{\thetable}{\textbf{b}}
% \normalsize
\caption{Comparisons of the robustness to Gaussian noise among different methods on Set11 dataset in the case of ratio = 10$\%$, where $^{\ast}$ represents the model is finetuned with mixed noises. And the difference of PSNR between w/o noise ($\sigma=0$) and w/ noise is highlighted in \textcolor{blue}{blue} color.}
\label{tab:sigma}
\begin{tabular}{ccc|cc|cc|cc}
% \hline 
\toprule
% \multicolumn{6}{c}{Set11 (PSNR/SSIM)} \\
% \hline
\multicolumn{1}{c|}{Noise}&\multicolumn{1}{c}{ISTA-Net$^{+}$\cite{zhang2018ista}}&\multicolumn{1}{c|}{ISTA-Net$^{+\ast}$}&\multicolumn{1}{c}{iPiano-Net\cite{su2020ipiano}}&\multicolumn{1}{c|}{iPiano-Net$^{\ast}$}&\multicolumn{1}{c}{DP-DUN}&\multicolumn{1}{c|}{DP-DUN$^{\ast}$}&\multicolumn{1}{c}{DPC-DUN}&\multicolumn{1}{c}{DPC-DUN$^{\ast}$}\\
\toprule
\multicolumn{1}{l|}{$\sigma=0$}&26.58(\ 0.00)&\multicolumn{1}{c|}{26.55(\ 0.00)}&28.33(\ 0.00)&28.15(\ 0.00)&29.42(\ 0.00)&\multicolumn{1}{c|}{29.23(\ 0.00)}&29.40(\ 0.00)&29.15(\ 0.00)\\
\multicolumn{1}{l|}{$\sigma=5$}&25.92(\textcolor{blue}{$\downarrow$0.66})&26.06(\textcolor{blue}{$\downarrow$0.49})&27.56(\textcolor{blue}{$\downarrow$0.77})&27.56(\textcolor{blue}{$\downarrow$0.59})&28.33(\textcolor{blue}{$\downarrow$1.09})&28.34(\textcolor{blue}{$\downarrow$0.89})&28.26(\textcolor{blue}{$\downarrow$1.14})&28.28(\textcolor{blue}{$\downarrow$0.87})\\
% \hline
\multicolumn{1}{l|}{$\sigma=7$}&25.43(\textcolor{blue}{$\downarrow$1.15})&25.70(\textcolor{blue}{$\downarrow$0.85})&27.01(\textcolor{blue}{$\downarrow$1.32})&27.15(\textcolor{blue}{$\downarrow$1.00})&27.60(\textcolor{blue}{$\downarrow$1.82})&27.80(\textcolor{blue}{$\downarrow$1.43})&27.58(\textcolor{blue}{$\downarrow$1.82})&27.76(\textcolor{blue}{$\downarrow$1.39})\\
% \hline
\multicolumn{1}{l|}{$\sigma=10$}&24.61(\textcolor{blue}{$\downarrow$1.97})&25.04(\textcolor{blue}{$\downarrow$1.51})&26.14(\textcolor{blue}{$\downarrow$2.19})&26.33(\textcolor{blue}{$\downarrow$1.82})&26.60(\textcolor{blue}{$\downarrow$2.82})&26.95(\textcolor{blue}{$\downarrow$2.28})&26.54(\textcolor{blue}{$\downarrow$2.86})&26.91(\textcolor{blue}{$\downarrow$2.24})\\
\toprule
\end{tabular}
\vspace{-0.2cm}
\end{table*}
%%%%%%%%%%%%%%%%%%%%%%%%%%%%%%%%%%%
%%%%%%%%%%%%%%%%%%%%%%%%%%%%%%%%%%%%%%%%
\begin{figure*}[!t]
\centering
% \vspace{-0.4cm}
\setlength{\abovecaptionskip}{0pt}
\setlength{\belowcaptionskip}{0cm}
\includegraphics[width=1.0\textwidth]{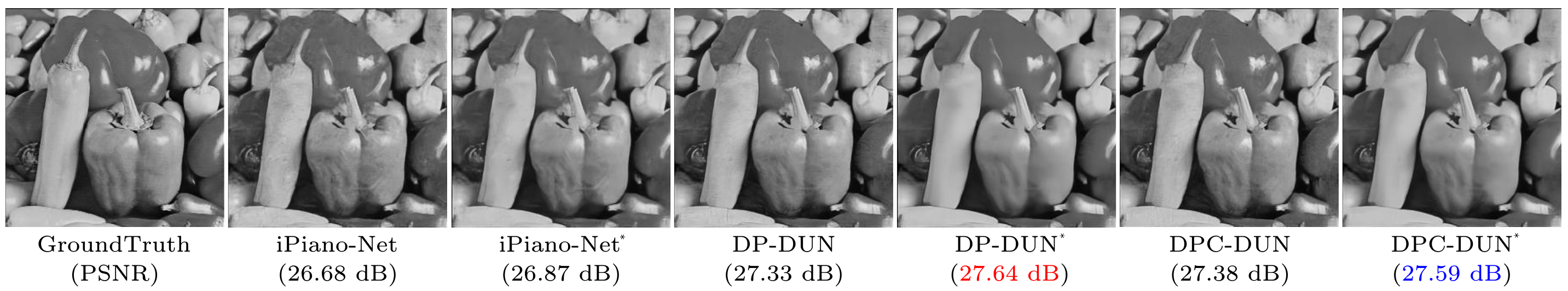}
% \vspace{-0.5cm}
\caption{Visualization of noise robustness on Set11 dataset \cite{Kulkarni2016ReconNetNR} when CS ratio = $10\%$ and $\sigma$ = 10, corresponding to Table~\ref{tab:sigma}. Our DP-DUN and DPC-DUN produce fewer stains and more accurate edges without artifacts.}
\vspace{-0.2cm}
\label{fig:noise}
\end{figure*}
%%%%%%%%%%%%%%%%%%%%%%%%%%%%%%%%%%%%%%%%
%%%%%%%%%%%%%%%%%%%%%%%%%%%%%%%%%%%%%%%%
\begin{figure*}[!t]
\centering
\setlength{\abovecaptionskip}{0.cm}
\setlength{\belowcaptionskip}{-0.cm}
\includegraphics[width=1.0\textwidth]{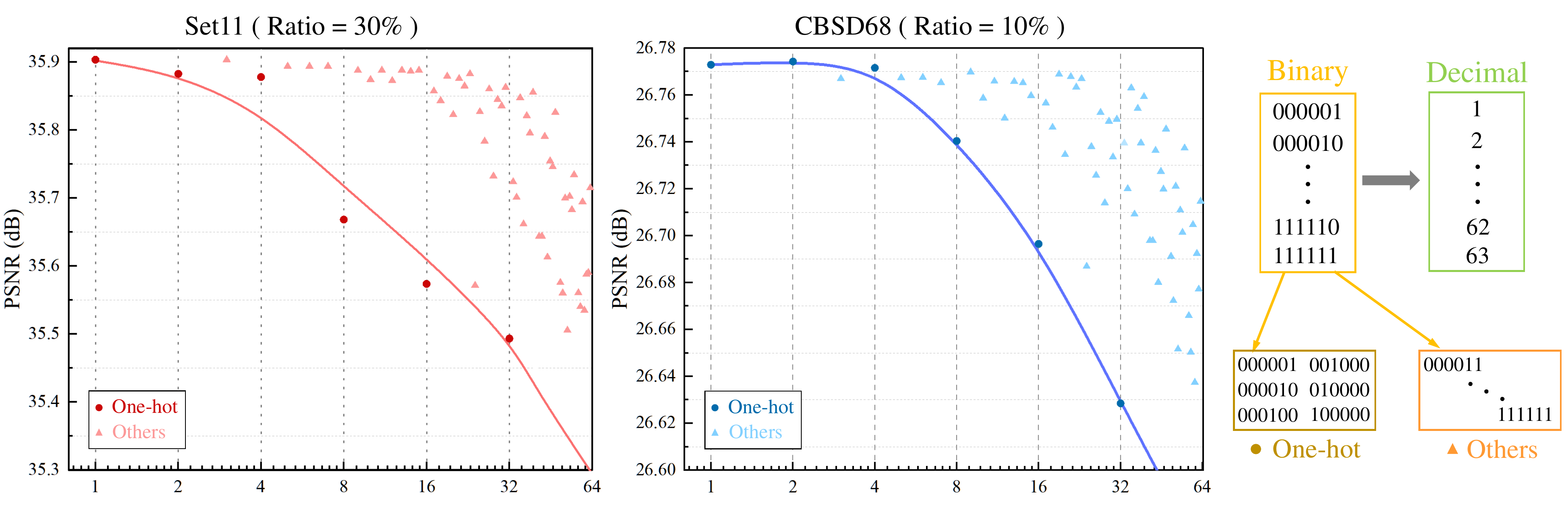} 
% \vspace{-0.4cm}
\caption{Scatter plots of the values in decimal format converted to binary and the corresponding PSNR values in two cases that are the same as Fig.~\ref{fig:eval_mu}.}
\vspace{-0.5cm}
\label{fig:seen}
\end{figure*}
%%%%%%%%%%%%%%%%%%%%%%%%%%%%%%%%%%%%%%%%

\subsubsection{Impact of Blocking Artifacts} To analyze the effect of image deblocking operations, we conduct the experiments of the models on Set11 dataset \cite{Kulkarni2016ReconNetNR} in Table~\ref{tab:deblock}, where ``-'' represents the results without finetuning on the patch size of $99\times99$. After finetuning, our models get PSNR/SSIM gains, but the number of the active module ($N_{AM}$) does not change obviously, indicating that deblocking can achieve a better performance-consumption balance.

\subsubsection{Application in the learnable sampling matrix} We use different sampling matrices to train our models to apply to different scenarios. We retrain DPC-DUN with the learnable matrix \cite{shi2019image} and compare it with some recent methods which also jointly learn the sampling matrix and the reconstruction process. We present the performance comparisons on Set11 dataset in Table~\ref{tab:learned_comp}. Our methods can be well applied to this matrix with high robustness and achieve good performance. 

\subsubsection{Sensitivity to Noise} In the real application, the imaging model may be affected by noise, so we add the experiments of the models with various Gaussian noises on Set11 dataset \cite{Kulkarni2016ReconNetNR} to demonstrate the robustness in Table~\ref{tab:sigma}. We finetune ISTA-Net$^+$ \cite{zhang2018ista}, iPiano-Net \cite{su2020ipiano}, DP-DUN, and DPC-DUN with the random noise in [0,10], namely ISTA-Net$^{+\ast}$, iPiano-Net$^{\ast}$, DP-DUN$^{\ast}$, and DPC-DUN$^{\ast}$ respectively. As shown in Table~\ref{tab:sigma}, our proposed models outperform the compared methods with all noises. The PSNR differences of DP-DUN$^{\ast}$/DPC-DUN$^{\ast}$ between w/o noise ($\sigma=0$) and w/ noise are smaller than DP-DUN/DPC-DUN, which indicates that the models after finetuning are relatively robust to different noises, especially larger ones in testing. Therefore, our proposed method can tackle the compressive sensing problem with the corresponding settings and datasets in real applications. What is more, Fig.~\ref{fig:noise} shows the visual results of noise robustness, from which one can see that the texture of the image is smoother and the noise is less when the models are finetuned.

\subsubsection{Analysis of Different Control Factors}

Our DPC-DUN can dynamically balance the recovery accuracy and complexity by combining a one-hot vector encoded by the control factor $\mu$ and the input image feature. In this paper, $\mu$ is set to $\{0.00001, 0.00005, 0.0001, 0.0005, 0.001, 0.002\}$ and corresponds to a one-hot vector $\{000001, 000010, 000100, 001000, \\ 010000, 100000\}$ as the input in the training phase. To break the limits of the six default one-hot vectors, we explore the effect of other (unseen) binary encoding vectors as the modulation during testing as shown in the lower row of Fig.~\ref{fig:eval_mu} and find that all vectors have similar trends. To explore the relationship between vectors and the modulation mechanism,
%of the performance-complexity tradeoff
we compare the decimal numbers converted to binary numbers and the corresponding PSNR values in two cases, as shown in Fig.~\ref{fig:seen}. Dots stand for the default settings, and triangles stand for the unseen settings. We can see that overall, the PSNR value decreases with the increase of the value in decimal format, and the default settings determine the extremum value of the reconstruction performance.

\section{Conclusion}\label{sec5}

We propose a \textbf{D}ynamic \textbf{P}ath-\textbf{C}ontrollable \textbf{D}eep \textbf{U}nfolding \textbf{N}etwork (\textbf{DPC-DUN}) for image CS, which shows a slimming mechanism that dynamically selects different paths and enables model interactivity with controllable complexity. Utilizing the skip connection structure inherent in DUNs, our proposed Path-Controllable Selector (PCS) can adaptively choose to skip a different number of modules for the different input images, thereby reducing the computational cost while ensuring similar high performance. And PCS also artificially adjusts the performance-complexity tradeoff at test time conditioned on the Lagrange multiplier. Our work opens new angles for the modulation of different tradeoffs in the image inverse tasks. However, our method also has some limitations. First, some background details may still be over-smoothed by our method. Second, our selector is not fine-grained enough to judge the picture information. Lastly, our selector adjustment range is relatively small and not continuous, which does not have a high modulation accuracy. In the future, we will improve our work by adjusting the sampling rate and extending our DPC-DUN to other network structures and video applications.
% \cite{9025255}.   

{
\bibliographystyle{IEEEtran}
\bibliography{egbib}

\begin{thebibliography}{10}
\providecommand{\url}[1]{#1}
\csname url@samestyle\endcsname
\providecommand{\newblock}{\relax}
\providecommand{\bibinfo}[2]{#2}
\providecommand{\BIBentrySTDinterwordspacing}{\spaceskip=0pt\relax}
\providecommand{\BIBentryALTinterwordstretchfactor}{4}
\providecommand{\BIBentryALTinterwordspacing}{\spaceskip=\fontdimen2\font plus
\BIBentryALTinterwordstretchfactor\fontdimen3\font minus
  \fontdimen4\font\relax}
\providecommand{\BIBforeignlanguage}[2]{{%
\expandafter\ifx\csname l@#1\endcsname\relax
\typeout{** WARNING: IEEEtran.bst: No hyphenation pattern has been}%
\typeout{** loaded for the language `#1'. Using the pattern for}%
\typeout{** the default language instead.}%
\else
\language=\csname l@#1\endcsname
\fi
#2}}
\providecommand{\BIBdecl}{\relax}
\BIBdecl

\bibitem{zhang2022bdu}
J.~Zhang, B.~Chen, R.~Xiong, and Y.~Zhang, ``Physics-inspired compressive
  sensing: Beyond deep unrolling,'' \emph{IEEE Signal Processing Magazine},
  vol.~40, no.~1, pp. 58--72, 2023.

\bibitem{candes2006robust}
E.~J. Cand{\`e}s, J.~Romberg, and T.~Tao, ``Robust uncertainty principles:
  Exact signal reconstruction from highly incomplete frequency information,''
  \emph{IEEE Transactions on information theory}, vol.~52, no.~2, pp. 489--509,
  2006.

\bibitem{sankaranarayanan2012cs}
A.~C. Sankaranarayanan, C.~Studer, and R.~G. Baraniuk, ``{CS-MUVI}: Video
  compressive sensing for spatial-multiplexing cameras,'' in \emph{Proceedings
  of the IEEE International Conference on Computational Photography (ICCP)},
  Apr. 2012.

\bibitem{liutkus2014imaging}
A.~Liutkus, D.~Martina, S.~Popoff, G.~Chardon, O.~Katz, G.~Lerosey, S.~Gigan,
  L.~Daudet, and I.~Carron, ``Imaging with nature: Compressive imaging using a
  multiply scattering medium,'' \emph{Scientific Reports}, vol.~4, p. 5552,
  2014.

\bibitem{zymnis2009compressed}
A.~Zymnis, S.~Boyd, and E.~Candes, ``Compressed sensing with quantized
  measurements,'' \emph{IEEE Signal Processing Letters}, vol.~17, no.~2, pp.
  149--152, 2009.

\bibitem{szczykutowicz2010dual}
T.~P. Szczykutowicz and G.~Chen, ``{Dual energy CT using slow kVp switching
  acquisition and prior image constrained compressed sensing},'' \emph{Physics
  in Medicine \& Biology}, vol.~55, no.~21, p. 6411, 2010.

\bibitem{zhang2022high}
J.~Zhang, Z.~Zhang, J.~Xie, and Y.~Zhang, ``High-throughput deep unfolding
  network for compressive sensing {MRI},'' \emph{IEEE Journal of Selected
  Topics in Signal Processing}, vol.~16, no.~4, pp. 750--761, 2022.

\bibitem{chen2019compressive}
Z.~Chen, X.~Hou, L.~Shao, C.~Gong, X.~Qian, Y.~Huang, and S.~Wang,
  ``Compressive sensing multi-layer residual coefficients for image coding,''
  \emph{IEEE Transactions on Circuits and Systems for Video Technology},
  vol.~30, no.~4, pp. 1109--1120, 2019.

\bibitem{duarte2008single}
M.~F. Duarte, M.~A. Davenport, D.~Takhar, J.~N. Laska, T.~Sun, K.~F. Kelly, and
  R.~G. Baraniuk, ``Single-pixel imaging via compressive sampling,'' \emph{IEEE
  Signal Processing Magazine}, vol.~25, no.~2, pp. 83--91, 2008.

\bibitem{rousset2016adaptive}
F.~Rousset, N.~Ducros, A.~Farina, G.~Valentini, C.~D’Andrea, and F.~Peyrin,
  ``Adaptive basis scan by wavelet prediction for single-pixel imaging,''
  \emph{IEEE Transactions on Computational Imaging}, vol.~3, no.~1, pp. 36--46,
  2016.

\bibitem{shen2013compressed}
H.~Shen, X.~Li, L.~Zhang, D.~Tao, and C.~Zeng, ``Compressed sensing-based
  inpainting of aqua moderate resolution imaging spectroradiometer band 6 using
  adaptive spectrum-weighted sparse {Bayesian} dictionary learning,''
  \emph{IEEE Transactions on Geoscience and Remote Sensing}, vol.~52, no.~2,
  pp. 894--906, 2013.

\bibitem{mou2022transcl}
C.~Mou and J.~Zhang, ``{TransCL}: Transformer makes strong and flexible
  compressive learning,'' \emph{IEEE Transactions on Pattern Analysis and
  Machine Intelligence}, vol.~45, no.~4, pp. 5236--5251, 2023.

\bibitem{wu2021spatial}
Z.~Wu, Z.~Zhang, J.~Song, and J.~Zhang, ``Spatial-temporal synergic prior
  driven unfolding network for snapshot compressive imaging,'' in
  \emph{Proceedings of IEEE International Conference on Multimedia and Expo
  (ICME)}, Jul. 2021.

\bibitem{wu2021ddun}
Z.~Wu, J.~Zhang, and C.~Mou, ``Dense deep unfolding network with {3D-CNN} prior
  for snapshot compressive sensing,'' in \emph{Proceedings of the IEEE
  International Conference on Computer Vision (ICCV)}, Oct. 2021, pp.
  4892--4901.

\bibitem{kim2010compressed}
Y.~Kim, M.~S. Nadar, and A.~Bilgin, ``Compressed sensing using a {Gaussian}
  scale mixtures model in wavelet domain,'' in \emph{Proceedings of the IEEE
  International Conference on Image Processing (ICIP)}, Sep 2010, pp.
  3365--3368.

\bibitem{Li2013AnEA}
C.~Li, W.~Yin, H.~Jiang, and Y.~Zhang, ``An efficient augmented lagrangian
  method with applications to total variation minimization,''
  \emph{Computational Optimization and Applications}, vol.~56, no.~3, pp.
  507--530, 2013.

\bibitem{zhang2014group}
J.~Zhang, D.~Zhao, and W.~Gao, ``Group-based sparse representation for image
  restoration,'' \emph{IEEE Transactions on Image Processing}, vol.~23, no.~8,
  pp. 3336--3351, 2014.

\bibitem{zhang2014image}
J.~Zhang, C.~Zhao, D.~Zhao, and W.~Gao, ``Image compressive sensing recovery
  using adaptively learned sparsifying basis via {L0} minimization,''
  \emph{Signal Processing}, vol. 103, pp. 114--126, 2014.

\bibitem{gao2015block}
X.~Gao, J.~Zhang, W.~Che, X.~Fan, and D.~Zhao, ``Block-based compressive
  sensing coding of natural images by local structural measurement matrix,'' in
  \emph{Proceedings of Data Compression Conference (DCC)}, Apr. 2015, pp.
  133--142.

\bibitem{Metzler2016FromDT}
C.~A. Metzler, A.~Maleki, and R.~G. Baraniuk, ``From denoising to compressed
  sensing,'' \emph{IEEE Transactions on Information Theory}, vol.~62, no.~9,
  pp. 5117--5144, 2016.

\bibitem{zhao2018cream}
C.~Zhao, J.~Zhang, R.~Wang, and W.~Gao, ``{CREAM}: {CNN-REgularized ADMM}
  framework for compressive-sensed image reconstruction,'' \emph{IEEE Access},
  vol.~6, pp. 76\,838--76\,853, 2018.

\bibitem{zhao2014image}
C.~Zhao, S.~Ma, and W.~Gao, ``Image compressive-sensing recovery using
  structured laplacian sparsity in {DCT} domain and multi-hypothesis
  prediction,'' in \emph{Proceedings of IEEE International Conference on
  Multimedia and Expo (ICME)}, Jul. 2014.

\bibitem{zhao2016video}
C.~Zhao, S.~Ma, J.~Zhang, R.~Xiong, and W.~Gao, ``Video compressive sensing
  reconstruction via reweighted residual sparsity,'' \emph{IEEE Transactions on
  Circuits and Systems for Video Technology}, vol.~27, no.~6, pp. 1182--1195,
  2016.

\bibitem{zhao2016nonconvex}
C.~Zhao, J.~Zhang, S.~Ma, and W.~Gao, ``Nonconvex {Lp} nuclear norm based
  {ADMM} framework for compressed sensing,'' in \emph{Proceedings of Data
  Compression Conference (DCC)}, Apr. 2016, pp. 161--170.

\bibitem{Kulkarni2016ReconNetNR}
K.~Kulkarni, S.~Lohit, P.~Turaga, R.~Kerviche, and A.~Ashok, ``{ReconNet}:
  Non-iterative reconstruction of images from compressively sensed
  measurements,'' in \emph{Proceedings of the IEEE Conference on Computer
  Vision and Pattern Recognition (CVPR)}, Jun. 2016, pp. 449--458.

\bibitem{sun2020dual}
Y.~Sun, J.~Chen, Q.~Liu, B.~Liu, and G.~Guo, ``Dual-path attention network for
  compressed sensing image reconstruction,'' \emph{IEEE Transactions on Image
  Processing}, vol.~29, pp. 9482--9495, 2020.

\bibitem{ren2021adaptive}
C.~Ren, X.~He, C.~Wang, and Z.~Zhao, ``Adaptive consistency prior based deep
  network for image denoising,'' in \emph{Proceedings of the IEEE Conference on
  Computer Vision and Pattern Recognition (CVPR)}, Jun. 2021, pp. 8596--8606.

\bibitem{zhang2018ista}
J.~Zhang and B.~Ghanem, ``{ISTA-Net}: Interpretable optimization-inspired deep
  network for image compressive sensing,'' in \emph{Proceedings of the IEEE
  Conference on Computer Vision and Pattern Recognition (CVPR)}, Jun. 2018, pp.
  1828--1837.

\bibitem{zhang2020optimization}
J.~Zhang, C.~Zhao, and W.~Gao, ``Optimization-inspired compact deep compressive
  sensing,'' \emph{IEEE Journal of Selected Topics in Signal Processing},
  vol.~14, no.~4, pp. 765--774, 2020.

\bibitem{you2021ista}
D.~You, J.~Xie, and J.~Zhang, ``{ISTA-Net$^{++}$}: Flexible deep unfolding
  network for compressive sensing,'' in \emph{Proceedings of IEEE International
  Conference on Multimedia and Expo (ICME)}, Jul. 2021.

\bibitem{you2021coast}
D.~You, J.~Zhang, J.~Xie, B.~Chen, and S.~Ma, ``{COAST}: Controllable
  arbitrary-sampling network for compressive sensing,'' \emph{IEEE Transactions
  on Image Processing}, vol.~30, pp. 6066--6080, 2021.

\bibitem{zhang2020amp}
Z.~Zhang, Y.~Liu, J.~Liu, F.~Wen, and C.~Zhu, ``{AMP-Net}: Denoising-based deep
  unfolding for compressive image sensing,'' \emph{IEEE Transactions on Image
  Processing}, vol.~30, pp. 1487--1500, 2020.

\bibitem{chen2016trainable}
Y.~Chen and T.~Pock, ``Trainable nonlinear reaction diffusion: A flexible
  framework for fast and effective image restoration,'' \emph{IEEE Transactions
  on Pattern Analysis and Machine Intelligence}, vol.~39, no.~6, pp.
  1256--1272, 2016.

\bibitem{lefkimmiatis2017non}
S.~Lefkimmiatis, ``Non-local color image denoising with convolutional neural
  networks,'' in \emph{Proceedings of the IEEE Conference on Computer Vision
  and Pattern Recognition (CVPR)}, Jul. 2017, pp. 3587--3596.

\bibitem{kruse2017learning}
J.~Kruse, C.~Rother, and U.~Schmidt, ``Learning to push the limits of efficient
  {FFT}-based image deconvolution,'' in \emph{Proceedings of the IEEE
  International Conference on Computer Vision (ICCV)}, Oct. 2017, pp.
  4596--4604.

\bibitem{wang2020stacking}
H.~Wang, T.~Zhang, M.~Yu, J.~Sun, W.~Ye, C.~Wang, and S.~Zhang, ``Stacking
  networks dynamically for image restoration based on the plug-and-play
  framework,'' in \emph{Proceedings of the European Conference on Computer
  Vision (ECCV)}, Aug. 2020, pp. 446--462.

\bibitem{kokkinos2018deep}
F.~Kokkinos and S.~Lefkimmiatis, ``Deep image demosaicking using a cascade of
  convolutional residual denoising networks,'' in \emph{Proceedings of the
  European Conference on Computer Vision (ECCV)}, Sep. 2018, pp. 317--333.

\bibitem{zhang2017learning}
K.~Zhang, W.~Zuo, S.~Gu, and L.~Zhang, ``Learning deep {CNN} denoiser prior for
  image restoration,'' in \emph{Proceedings of the IEEE Conference on Computer
  Vision and Pattern Recognition (CVPR)}, Jul. 2017, pp. 3929--3938.

\bibitem{dong2018denoising}
W.~Dong, P.~Wang, W.~Yin, G.~Shi, F.~Wu, and X.~Lu, ``Denoising prior driven
  deep neural network for image restoration,'' \emph{IEEE Transactions on
  Pattern Analysis and Machine Intelligence}, vol.~41, no.~10, pp. 2305--2318,
  2018.

\bibitem{gilton2019neumann}
D.~Gilton, G.~Ongie, and R.~Willett, ``Neumann networks for linear inverse
  problems in imaging,'' \emph{IEEE Transactions on Computational Imaging},
  vol.~6, pp. 328--343, 2019.

\bibitem{song2021memory}
J.~Song, B.~Chen, and J.~Zhang, ``Memory-augmented deep unfolding network for
  compressive sensing,'' in \emph{Proceedings of the ACM International
  Conference on Multimedia (ACM MM)}, Oct. 2021, pp. 4249--4258.

\bibitem{chenlearning}
J.~Chen, Y.~Sun, Q.~Liu, and R.~Huang, ``Learning memory augmented cascading
  network for compressed sensing of images,'' in \emph{Proceedings of the
  European Conference on Computer Vision (ECCV)}, Aug. 2020, pp. 513--529.

\bibitem{su2020ipiano}
Y.~Su and Q.~Lian, ``{iPiano-Net}: Nonconvex optimization inspired multi-scale
  reconstruction network for compressed sensing,'' \emph{Signal Processing:
  Image Communication}, vol.~89, p. 115989, 2020.

\bibitem{yu2021path}
K.~Yu, X.~Wang, C.~Dong, X.~Tang, and C.~C. Loy, ``{Path-Restore}: Learning
  network path selection for image restoration,'' \emph{IEEE Transactions on
  Pattern Analysis and Machine Intelligence}, vol.~44, no.~10, pp. 7078--7092,
  2021.

\bibitem{han2021dynamic}
Y.~Han, G.~Huang, S.~Song, L.~Yang, H.~Wang, and Y.~Wang, ``Dynamic neural
  networks: A survey,'' \emph{IEEE Transactions on Pattern Analysis and Machine
  Intelligence}, vol.~44, no.~11, pp. 7436--7456, 2021.

\bibitem{zhu2021dynamic}
M.~Zhu, K.~Han, E.~Wu, Q.~Zhang, Y.~Nie, Z.~Lan, and Y.~Wang, ``Dynamic
  resolution network,'' in \emph{Proceedings of the International Conference on
  Neural Information Processing Systems (NeurIPS)}, Dec. 2021.

\bibitem{li2021dynamic}
C.~Li, G.~Wang, B.~Wang, X.~Liang, Z.~Li, and X.~Chang, ``Dynamic slimmable
  network,'' in \emph{Proceedings of the IEEE Conference on Computer Vision and
  Pattern Recognition (CVPR)}, Jun. 2021, pp. 8607--8617.

\bibitem{wu2018blockdrop}
Z.~Wu, T.~Nagarajan, A.~Kumar, S.~Rennie, L.~S. Davis, K.~Grauman, and
  R.~Feris, ``{BlockDrop}: Dynamic inference paths in residual networks,'' in
  \emph{Proceedings of the IEEE Conference on Computer Vision and Pattern
  Recognition (CVPR)}, Jun. 2018, pp. 8817--8826.

\bibitem{wang2018skipnet}
X.~Wang, F.~Yu, Z.-Y. Dou, T.~Darrell, and J.~E. Gonzalez, ``{SkipNet}:
  Learning dynamic routing in convolutional networks,'' in \emph{Proceedings of
  the European Conference on Computer Vision (ECCV)}, Sep. 2018, pp. 420--436.

\bibitem{song2019dynamic}
Y.~Song, Y.~Zhu, and X.~Du, ``Dynamic residual dense network for image
  denoising,'' \emph{Sensors}, vol.~19, no.~17, p. 3809, 2019.

\bibitem{he2016deep}
K.~He, X.~Zhang, S.~Ren, and J.~Sun, ``Deep residual learning for image
  recognition,'' in \emph{Proceedings of the IEEE Conference on Computer Vision
  and Pattern Recognition (CVPR)}, Jun. 2016, pp. 770--778.

\bibitem{he2021interactive}
J.~He, C.~Dong, Y.~Liu, and Y.~Qiao, ``Interactive multi-dimension modulation
  for image restoration,'' \emph{IEEE Transactions on Pattern Analysis and
  Machine Intelligence}, vol.~44, no.~12, pp. 9363--9379, 2021.

\bibitem{cai2021toward}
H.~Cai, J.~He, Y.~Qiao, and C.~Dong, ``Toward interactive modulation for
  photo-realistic image restoration,'' in \emph{Proceedings of the IEEE
  Conference on Computer Vision and Pattern Recognition Workshops (CVPRW)},
  Jun. 2021.

\bibitem{jiang2021towards}
J.~Jiang, K.~Zhang, and R.~Timofte, ``Towards flexible blind {JPEG} artifacts
  removal,'' in \emph{Proceedings of the IEEE International Conference on
  Computer Vision (ICCV)}, Oct. 2021, pp. 4977--4986.

\bibitem{choi2019variable}
Y.~Choi, M.~El-Khamy, and J.~Lee, ``Variable rate deep image compression with a
  conditional autoencoder,'' in \emph{Proceedings of the IEEE Conference on
  Computer Vision and Pattern Recognition (CVPR)}, Jun. 2019, pp. 3146--3154.

\bibitem{lin2020variable}
J.~Lin, M.~Akbari, H.~Fu, Q.~Zhang, S.~Wang, J.~Liang, D.~Liu, F.~Liang,
  G.~Zhang, and C.~Tu, ``Variable-rate multi-frequency image compression using
  modulated generalized octave convolution,'' in \emph{Proceedings of the
  International Workshop on Multimedia Signal Processing (MMSP)}, Sep. 2020.

\bibitem{jang2016categorical}
E.~Jang, S.~Gu, and B.~Poole, ``Categorical reparameterization with
  {Gumbel-Softmax},'' in \emph{Proceedings of the International Conference on
  Learning Representations (ICLR)}, Apr. 2017.

\bibitem{dai2021mix}
T.~Dai, Y.~Lv, B.~Chen, Z.~Wang, Z.~Zhu, and S.-T. Xia, ``Mix-order attention
  networks for image restoration,'' in \emph{Proceedings of the ACM
  International Conference on Multimedia (ACM MM)}, Oct. 2021, pp. 2880--2888.

\bibitem{hu2018squeeze}
J.~Hu, L.~Shen, and G.~Sun, ``Squeeze-and-excitation networks,'' in
  \emph{Proceedings of the IEEE Conference on Computer Vision and Pattern
  Recognition (CVPR)}, Jun. 2018, pp. 7132--7141.

\bibitem{zhang2021plug}
K.~Zhang, Y.~Li, W.~Zuo, L.~Zhang, L.~Van~Gool, and R.~Timofte, ``Plug-and-play
  image restoration with deep denoiser prior,'' \emph{IEEE Transactions on
  Pattern Analysis and Machine Intelligence}, vol.~44, no.~10, pp. 6360--6376,
  2021.

\bibitem{Agustsson_2017_CVPR_Workshops}
E.~Agustsson and R.~Timofte, ``{NTIRE} 2017 challenge on single image
  super-resolution: Dataset and study,'' in \emph{Proceedings of the IEEE
  Conference on Computer Vision and Pattern Recognition Workshops (CVPRW)},
  Jul. 2017.

\bibitem{zhong2022scalable}
Y.~Zhong, C.~Zhang, F.~Ren, H.~Kuang, and P.~Tang, ``Scalable image compressed
  sensing with generator networks,'' \emph{IEEE Transactions on Computational
  Imaging}, vol.~8, pp. 1025--1037, 2022.

\bibitem{ma2022deep}
C.~Ma, J.~T. Zhou, X.~Zhang, and Y.~Zhou, ``Deep unfolding for compressed
  sensing with denoiser,'' in \emph{Proceedings of IEEE International
  Conference on Multimedia and Expo (ICME)}, Jul. 2022.

\bibitem{shen2022transcs}
M.~Shen, H.~Gan, C.~Ning, Y.~Hua, and T.~Zhang, ``{TransCS}: A
  transformer-based hybrid architecture for image compressed sensing,''
  \emph{IEEE Transactions on Image Processing}, vol.~31, pp. 6991--7005, 2022.

\bibitem{shi2019image}
W.~Shi, F.~Jiang, S.~Liu, and D.~Zhao, ``Image compressed sensing using
  convolutional neural network,'' \emph{IEEE Transactions on Image Processing},
  vol.~29, pp. 375--388, 2019.

\end{thebibliography}
}
%\begin{biography}[{\includegraphics[width=1in,height=1.25in,clip,keepaspectratio]{mshell}}]{Michael Shell}


%\vfill


% \subsection{Comparison with State-of-the-Art Methods}
% To determine a proper residual block number $N_p$, we plot the average PSNR curves by OPTICS for Set11 with respect to different Residual Block Number in the cases of CS ratio=30\%, as shown in Fig.~\ref{fig: phase_loss}(a).
% One can observe that the PSNR increase as residual block number $N_p$ increases; however, the model with 3 and 4 $N_p$ have similar performance. Thus, considering the trade-off between computational complexity and recover performance, we set residual block number to be 3 for our OPTICS by default. 

\vspace{-1.2cm}
\begin{IEEEbiography}
[\vspace{-0.8cm}{\includegraphics[width=1in,height=1.25in,clip,keepaspectratio]{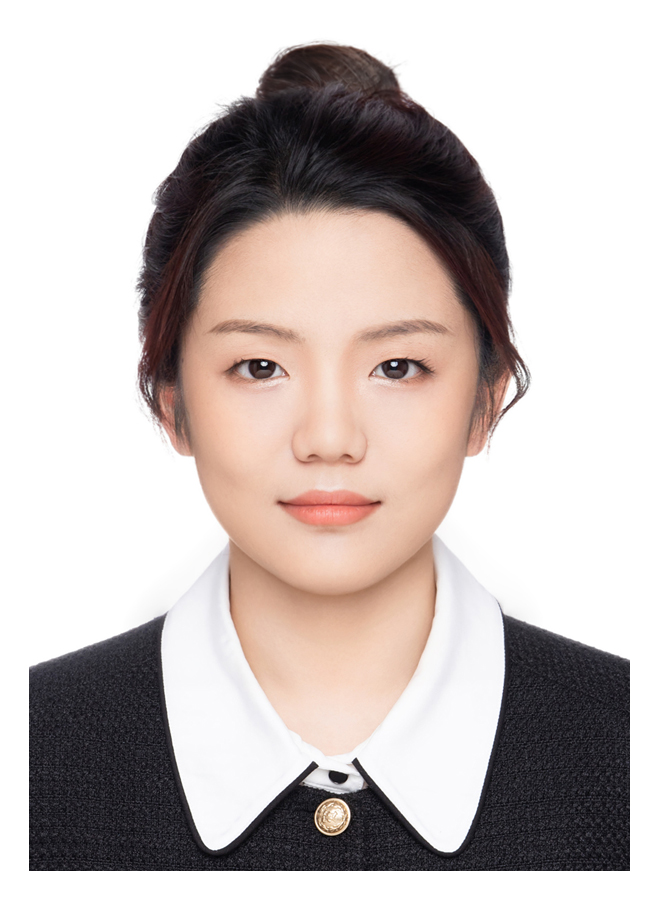}}]{Jiechong Song} received the B.E. degree in the School of Electronics and Information Engineering, Sichuan University, Chengdu, China, in 2019. She is currently working toward a doctor's degree in computer applications technology at Peking University Shenzhen Graduate School, Shenzhen, China. Her research interests include compressive sensing, image restoration, and computer vision. 

\end{IEEEbiography}

\vspace{-1.2cm}
\begin{IEEEbiography}
[\vspace{-0.8cm}{\includegraphics[width=1in,height=1.25in,clip,keepaspectratio]{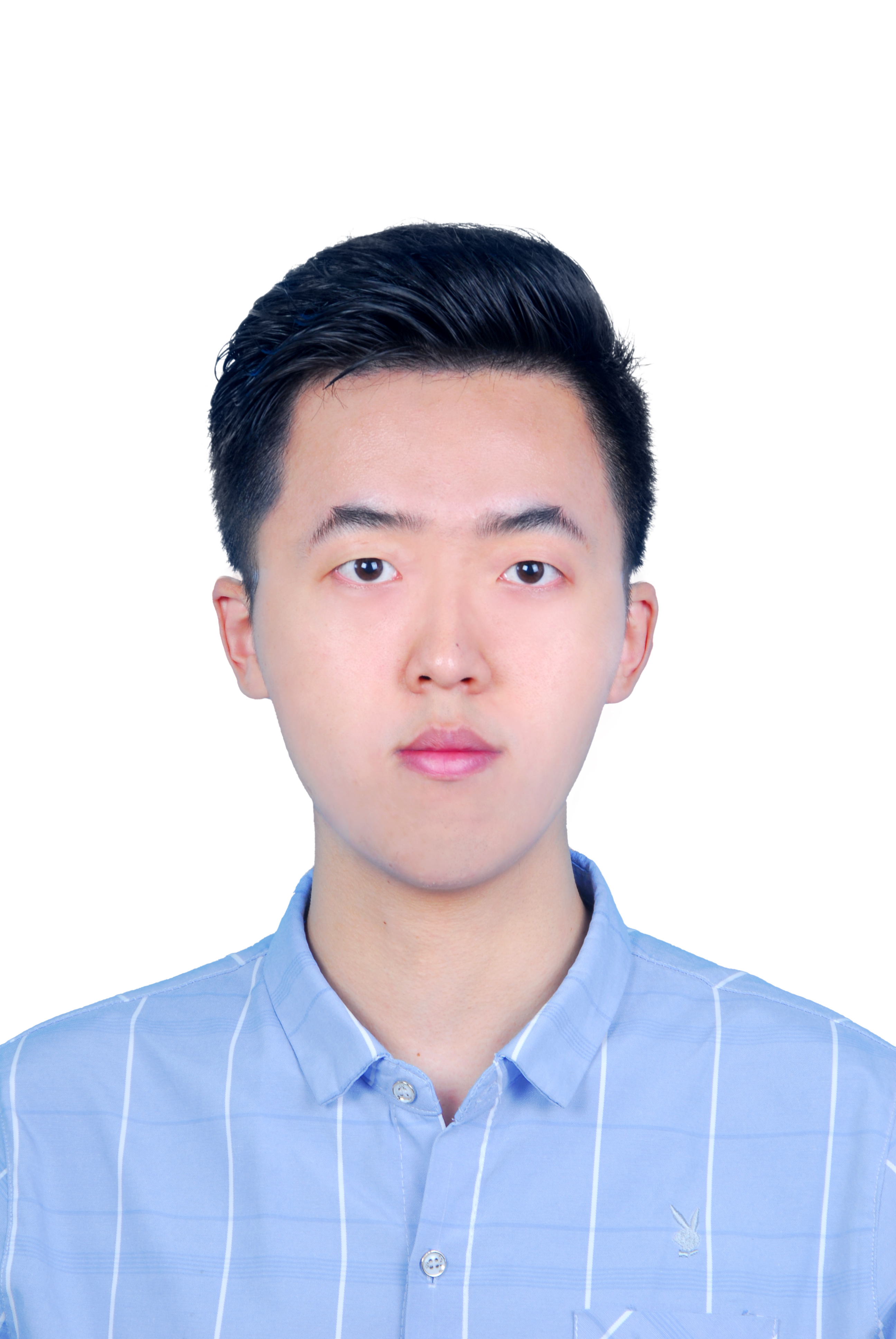}}]{Bin Chen} received the B.E. degree in the School of Computer Science, Beijing University of Posts and Telecommunications, Beijing, China, in 2021. He is currently working toward the master's degree in computer applications technology at Peking University Shenzhen Graduate School, Shenzhen, China. His research interests include compressive sensing, image restoration and computer vision. 

\end{IEEEbiography}

\vspace{-1.2cm}
\begin{IEEEbiography}[{\includegraphics[width=1in,height=1.25in,clip,keepaspectratio]{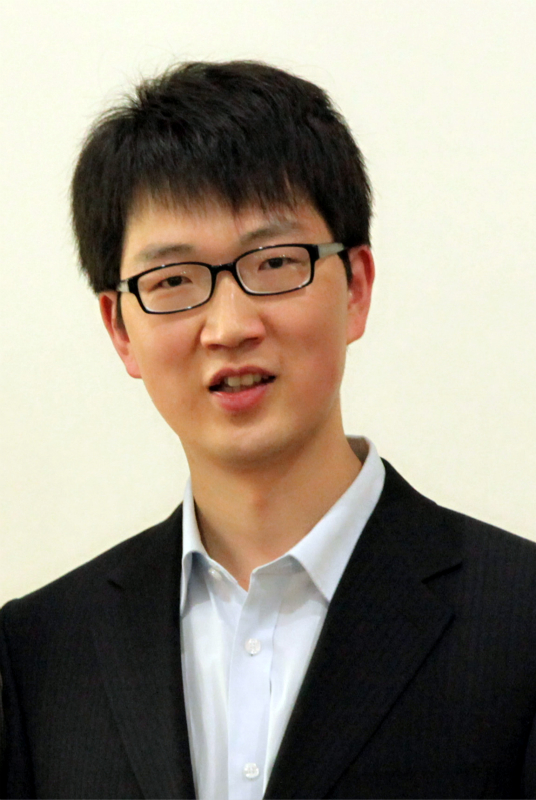}}]{Jian Zhang} (M'14) received the B.S. degree from the Department of Mathematics, Harbin Institute of Technology (HIT), Harbin, China, in 2007, and received his M.Eng. and Ph.D. degrees from the School of Computer Science and Technology, HIT, in 2009 and 2014, respectively. From 2014 to 2018, he worked as a postdoctoral researcher at Peking University (PKU), Hong Kong University of Science and Technology (HKUST), and King Abdullah University of Science and Technology (KAUST). 

Currently, he is an Assistant Professor with the School of Electronic and Computer Engineering, Peking University Shenzhen Graduate School, Shenzhen, China. His research interests include intelligent multimedia processing, low-level vision and computational imaging. He is now leading the Visual-Information Intelligent Learning LAB (VILLA) at PKU. He has published over 90 technical articles in refereed international journals and proceedings, including SPM/TPAMI/TIP/CVPR/ECCV/ICCV/ICLR. He received the Best Paper Award at the 2011 IEEE Visual Communications and Image Processing (VCIP) and was a co-recipient of the Best Paper Award of 2018 IEEE MultiMedia. 
\end{IEEEbiography}

\end{document}